\documentclass{article}

\usepackage{PRIMEarxiv}

\usepackage[utf8]{inputenc} 
\usepackage[T1]{fontenc}    

\usepackage{graphicx} 
\usepackage{amsmath} 
\usepackage{amssymb} 
\usepackage{xcolor}
\usepackage{booktabs}
\usepackage{url}
\usepackage{pdflscape}
\usepackage{float}
\usepackage{multirow}
\usepackage{booktabs}
\usepackage{array}
\usepackage{algorithm} 
\usepackage{algorithmic}
\usepackage{amsmath}
\usepackage{placeins}
\usepackage{hyperref}
\usepackage{pdflscape}
\usepackage{afterpage}
\usepackage{caption}
\usepackage[table]{xcolor}
\usepackage[numbers]{natbib}

\title{Adaptive Operator Selection in Bilevel Large
Neighborhood Search for Electric Autonomous
Dial-a-Ride Problem under Uncertainty}

\author{ \href{https://orcid.org/0000-0002-8043-994X}{\includegraphics[scale=0.06]{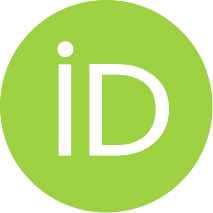}\hspace{1mm}Ishara Hewa Pathiranage} \\
	Machine Learning and Optimisation, \\School of Computer Science and \\Information Technology,\\
  Adelaide University,\\ Adelaide, Australia 
	\And
	\href{https://orcid.org/0000-0002-0036-4782}{\includegraphics[scale=0.06]{orcid.pdf}\hspace{1mm}Aneta Neumann} \\
	Machine Learning and Optimisation, \\
    School of Computer Science and\\ Information Technology,\\
  Adelaide University,\\ 
  Adelaide, Australia \\
}

\begin{document}

\maketitle

\begin{abstract}
The electric autonomous dial-a-ride problem (EADARP) extends the classical dial-a-ride problem by incorporating battery and charging constraints for electric vehicles. In practice, travel-time uncertainty can cause violations of time-window constraints. Large neighborhood search is effective for solving the EADARP, but its performance can depend on the choice of insertion operator during the repair phase. This paper investigates insertion-operator selection within a bilevel large neighborhood search framework for deterministic and chance-constrained variants of the EADARP. In the chance-constrained variant, arc travel times are modeled as independent normally distributed random variables, and upper time-window constraints are enforced probabilistically. We consider six selection methods, namely fixed greedy insertion, fixed regret-based insertion, random selection, a deterministic state-based rule, performance-adaptive ALNS selection, and LLM-based state-aware selection. Experimental results show comparable performance on smaller instances, while differences become more evident on larger and more constrained instances. There is no single strategy that performs best across all instances, and the relative performance of the LLM-based, rule-based, and ALNS strategies varies with the problem instance and experimental setting.
\end{abstract}


\section{Introduction}

The growing demand for urban transportation has increased interest in on-demand ride-sharing systems~\cite{SHAHEEN202037}. Compared with fixed-schedule public transportation, these services offer greater flexibility. They may also reduce operating costs and environmental impacts relative to privately owned vehicles and conventional taxi services. Recent advances in electric and autonomous vehicle technologies enable energy-efficient and scalable operations. Electric vehicles introduce constraints related to battery capacity and charging, while autonomous vehicles enable continuous operation without human drivers and improve fleet utilization. The integration of these technologies leads to the electric autonomous dial-a-ride problem (EADARP)~\cite{NARAYANAN2020255}, which extends the classical dial-a-ride problem (DARP)~\cite{Molenbruch2017,HO2018395} by incorporating battery and charging constraints. These additional constraints further increase the complexity of the already NP-hard DARP. 

Moreover, travel times in ride-sharing networks can vary due to traffic congestion, incidents, weather, and time-of-day effects. Ignoring such uncertainty may lead to time-window violations, particularly in tightly scheduled services with limited time buffer. To address this problem, chance constraints~\cite{Charnes} can be incorporated into optimization problems such that stochastic constraints are satisfied with high probability, usually denoted by~$\alpha$. It has been successfully applied to optimization problems in domains such as mining~\cite{doi:10.1080/25726668.2021.1916170,10.1145/3449639.3459382,10254112} and logistics~\cite{10.5555/2888116.2888219,
FARINA201653,JOHNSEN2025103968}.

Large neighborhood search (LNS) is a well-established local search heuristic that uses a destroy operator to remove part of the current solution and a repair operator to reconstruct the resulting partial solution at each iteration~\cite{Pisinger2010}. In the literature, a variety of operators have been proposed for request removal and insertion. Common removal operators include random and worst removal, while popular insertion operators include greedy, deep-greedy, and regret-based insertion~\cite{Ropke2006,10.1007/978-3-319-68496-3_5}. LNS is an established method to solve DARP~\cite{BRAEKERS2014166,MASMOUDI201660,MOLENBRUCH201797},  EADARP~\cite{LIMMER2023100876,10.1145/3638529.3654057,10.1007/978-3-031-82949-9_20,10.1145/3779426}, and other routing, scheduling, and location problems~\cite{10.1007/978-3-030-94216-8_1,10.1007/3-540-49481-2_30}.

Furthermore, recent advances in artificial intelligence and natural language processing have led to the rapid development of large language models (LLMs). These models have been successfully applied across a wide range of applications, including decision support systems~\cite{smartcities8010019,HANDLER2024102811,ZHONG2025103042} and tools for algorithm design~\cite{10.1145/3787585,10.1007/978-981-96-0348-0_7,10.5555/3737916.3739297}.

In this paper, we investigate insertion-operator selection within a bilevel large neighborhood search framework for deterministic and chance-constrained EADARP variants. We compare fixed greedy, fixed regret, random, deterministic rule-based, performance-adaptive ALNS, and LLM-based state-aware selection while keeping all other components of the BI-LNS framework unchanged.

\subsection{Related Work}

Several exact and metaheuristic approaches have been proposed for solving the EADARP. Bongiovanni et al.~\cite{BONGIOVANNI2019436} introduced the problem and proposed three- and two-index MILP formulations, demonstrating improved scalability of the two-index model. Subsequently, Bongiovanni et al.~\cite{BONGIOVANNI2022102835} investigated the dynamic EADARP using a greedy construction combined with a machine learning–assisted large neighborhood search. Due to the limited scalability of exact methods, several metaheuristic approaches have been developed, including deterministic annealing with battery-restricted fragments~\cite{SU20231091} and the bilevel large neighborhood search (BI-LNS) framework~\cite{LIMMER2023100876}. In BI-LNS, an outer level optimizes charging schedules, while an inner level optimizes pickup and delivery requests. More recent studies further extended LNS-based approaches with dynamic charging-station insertion strategies~\cite{10.1145/3638529.3654057,10.1007/978-3-031-82949-9_20,10.1145/3779426,10.1145/3795095.3805133,10.1145/3638529.3654067}. Moreover, early work by Ropke
and Pisinger~\cite{Ropke2006} examined operator selection in LNS, proposing adaptive mechanisms that choose among multiple heuristics using historical performance information.

Chance-constrained optimization~\cite{Charnes} provides a promising framework for handling uncertainty by ensuring that constraints are satisfied with a specified probability~\cite{DBLP:conf/gecco/0001W23,10.1145/3638529.3654066,10.1162/evco_a_00360}. In the mining domain, Stimson et al.~\cite{10254112} applied a single-objective evolutionary algorithm to chance-constrained open-pit mine scheduling. More recently, a bi-objective evolutionary formulation was developed for dynamic chance-constrained open-pit mine scheduling, where block economic values are stochastic and mining and processing capacities vary over time~\cite{DynamicOPMSP}.
In the context of DARP, Johnsen et al.~\cite{JOHNSEN2025103968} investigated a stochastic variant with interrelated trips. They proposed a chance-constrained formulation that approximates arrival- and service-time distributions and evaluates probabilistic feasibility within an adaptive variable neighborhood search framework.

Recent studies have also explored LLMs as tools for decision support and automated algorithm design in optimization. Zhong et al.~\cite{ZHONG2025103042} integrated LLMs into hyper-heuristic evolutionary frameworks, where the LLM acts as a high-level controller for selecting or sequencing low-level heuristics. Liu et al.~\cite{10.1145/3787585} provided a systematic review outlining the roles and limitations of LLMs across the algorithm design pipeline. In addition, language-driven hyper-heuristics have been proposed for NP-hard combinatorial optimization, enabling automated heuristic generation and adaptation with minimal manual intervention~\cite{10.5555/3737916.3739297}.

Despite these advances, most EADARP studies consider deterministic travel times, and the effect of insertion-operator selection within BI-LNS remains underexplored. In particular, fixed, random, rule-based, and adaptive selection strategies have not been compared for deterministic and chance-constrained EADARP variants.

\subsection{Contribution}

This paper investigates inner-level insertion-operator selection within BI-LNS for classical and chance-constrained variants of the EADARP. Building on the BI-LNS framework introduced in~\cite{LIMMER2023100876}, we study how different selection strategies affect feasibility and solution quality while keeping the remaining components of the algorithm unchanged.

We compare six insertion-operator selection strategies, including greedy insertion, regret-based insertion, random selection, deterministic rule-based selection, adaptive ALNS selection, and LLM-based state-aware selection. We introduce BI-LNS-LLM, in which an LLM selects between greedy and regret-based insertion using compact features that describe the current search state. The LLM does not construct routes, check feasibility, or evaluate objective values. 

We also consider a chance-constrained EADARP in which arc travel times are modeled as independent, normally distributed random variables, while all other problem parameters remain deterministic. Upper time-window constraints are enforced probabilistically using chance constraints. 

The remainder of this paper presents the deterministic and chance-constrained EADARP formulations in Sections~\ref{sec:problem_definition} and~\ref{sec:cc_eadarp}, respectively. Section~\ref{sec:method} describes the BI-LNS framework and operator-selection strategies. Sections~\ref{sec:experimental_setting} and~\ref{sec:experimental_results} present the experimental setup and results, respectively, and Section~\ref{sec:conclusion} concludes the paper.

\section{Problem Definition}
\label{sec:problem_definition}

This section defines the electric autonomous dial-a-ride problem, including its objective function and constraints.

\subsection{Problem Setting}

The EADARP extends the classical DARP by incorporating battery and charging constraints for electric autonomous vehicles (EAVs). The problem considered in this work is based on the deterministic EADARP variant of Limmer~\cite{LIMMER2023100876},
itself derived from the formulation introduced by Bongiovanni et al.~\cite{BONGIOVANNI2019436}. We define the EADARP as a complete directed graph $G=(V,A)$, where $V$ denotes the set of locations and $A=\{(i,j)\mid i,j\in V,\ i\neq j\}$ denotes the set of arcs. A fleet of $K$ electric autonomous vehicles is available to serve $N$ transportation requests over a planning horizon of length $T$.

Let $\mathcal{K}=\{1,\dots,K\}$ denote the set of vehicles. The vertex set $V$ is partitioned into five disjoint subsets: the set of pickup locations $P=\{1,\dots,N\}$, the corresponding set of drop-off locations $D=\{N+1,\dots,2N\}$,
the set of start depots $O$, the set of end depots $F$, and the set of charging stations $S$. Each transportation request $i\in P$ consists of a pickup at location $i$ and a corresponding drop-off at location $i+N\in D$. Each vehicle $k\in\mathcal{K}$ starts its route at a designated start depot $o_k\in O$ and must terminate at an end depot $f_k\in F$.

\subsection{Travel Times, Time Windows, and Loads}

Each arc $(i,j)\in A$ has a travel time $t_{i,j}$, defined as the Euclidean distance between the corresponding locations, and an energy consumption $e_{i,j} = \beta\, t_{i,j}$, where $\beta$ is the energy consumption rate per unit travel time.
Each location $i\in V$ has a time window $[w_i^{\text{start}}, w_i^{\text{end}}]$ within which service must begin.
Service at each pickup and drop-off location $i \in P \cup D$ requires a positive duration $d_i > 0$, while $d_i = 0$ for depots and charging stations. Let $a_i$ and $s_i$ denote the arrival time and the service start time at location $i$,
respectively.

Vehicle loads are defined by $l_i$, $l_i > 0$ for pickup locations, $l_{i+N} = -l_i$ for the corresponding drop-off locations, and $l_i = 0$ for all other locations. Each vehicle $k\in\mathcal{K}$ has a maximum load capacity $C_k$.

\subsection{Battery and Charging Constraints}

Each vehicle $k \in \mathcal{K}$ has battery capacity $Q$ and starts its route with an initial battery level $Q_k^0$. The battery level upon arrival at the destination depot must satisfy $Q_k^{\text{end}} \ge \gamma Q$, where $\gamma \in [0,1]$.
Each charging station $s \in S$ has a charging rate $\rho_s$. Vehicles may partially recharge when visiting charging stations. Let $g_{k,h}$ denote the charging gain obtained by vehicle $k$ at its $h$-th visited location, which is zero if the location is not a charging station. Note that only vehicles without passengers may visit charging stations. In this work, we consider a relaxed EADARP variant in which charging stations may be visited an unlimited number of times, as also assumed in~\cite{BONGIOVANNI2019436, LIMMER2023100876, 10.1145/3638529.3654057}. We model this by allowing charging station locations to appear multiple times within a route, while all other locations may be visited at most once.

\subsection{Routes and Feasibility}

A route for a vehicle $k \in \mathcal{K}$ is defined as an ordered sequence of locations that starts at $o_k\in O$, may visit pickup, drop-off, and charging station locations, and ends at $f_k\in F$. A solution to the EADARP consists of one route per vehicle together with the service start times at all visited locations.

A solution is feasible if the following conditions are satisfied. Each request is served exactly once. For every request $i \in P$, exactly one route visits the pickup location $i$ and the corresponding drop-off location $i+N$, with the pickup occurring before the drop-off. The load of each vehicle $k \in \mathcal{K}$ never exceeds its capacity $C_k$, and the maximum ride time constraint is satisfied for every request. Service at each location $i \in V$ starts within its time window
$[w_i^{\text{start}}, w_i^{\text{end}}]$; vehicles can arrive before $w_i^{\text{start}}$ and wait, but service may not start after $w_i^{\text{end}}$.

Each vehicle starts its route with battery level $Q_k^0$, never depletes its battery, never charges beyond its battery capacity $Q$, and satisfies the minimum final battery requirement upon arrival at its end depot. Vehicles may visit charging stations only when they carry no passengers. A physical charging station may be visited multiple times, whereas each pickup and drop-off location may be visited only once.

\subsection{Objective Function}

The objective of the EADARP is to find a feasible solution that minimizes a weighted sum of total travel time across vehicle routes and total passenger excess ride time for all transportation requests, $i \in P$, i.e.,

\begin{equation}
\label{eq:objective}
\min \;
w_1 \sum_{k \in \mathcal{K}} \sum_{(i,j)\in A} t_{i,j} \, x_{i,j}^k
\;+\;
w_2 \sum_{i \in P} r_i .
\end{equation}
Here, $w_1$ and $w_2$ denote the weight factors associated with routing cost and passenger inconvenience, respectively, and $x_{i,j}^k \in \{0,1\}$ is a binary decision variable that equals $1$ if vehicle $k$ travels directly from location $i$ to
$j$, and $0$ otherwise.

The excess ride time $r_i$ of request $i \in P$ measures the additional delay experienced by a passenger relative to a direct trip from pickup to drop-off and is defined as $r_i = s_{i+N} - s_i - d_i - t_{i,i+N}$, where $s_i$ and $s_{i+N}$ denote the service start times at the pickup and drop-off locations, respectively, $d_i$ is the service duration at the pickup location, and $t_{i,i+N}$ denotes the direct travel time from pickup location $i$ to its corresponding drop-off location $i+N$.

\section{Chance-Constrained EADARP with Stochastic Travel Times}
\label{sec:cc_eadarp}

We extend the deterministic EADARP by incorporating travel-time uncertainty while keeping all remaining problem parameters deterministic. Travel-time uncertainty propagates along vehicle routes through stochastic travel times and deterministic service and charging durations.

For each arc $(i,j)\in A$, we model the travel time as an independent normally distributed random variable $\tilde t_{i,j} \sim \mathcal{N}(t_{i,j}, \sigma_{i,j}^2)$, where $t_{i,j}$ denotes the mean travel time and $\sigma_{i,j}=\mathrm{CV}\cdot t_{i,j}$ such that $\mathrm{CV}\geq 0$ is the coefficient of variation controlling the level of travel-time uncertainty. We assume that travel times on different arcs are mutually independent. Consider a route of vehicle $k\in\mathcal{K}$ visiting the ordered sequence $(v_0^k,v_1^k,\ldots,v_{L_k}^k)$, where $v_0^k=o_k$ denotes its start depot. Let $\tilde a_{v_\ell^k}$ denote the random arrival time at location $v_\ell^k$. Because service cannot begin before the lower bound of the corresponding time window, the random service-start time is $\tilde s_i = \max\!\big(\tilde a_i,\, w_i^{\text{start}}\big)$.

Let $q_{v_\ell^k}$ denote the effective deterministic charging duration used in the stochastic time propagation when $v_\ell^k$ is a charging station, and let $q_{v_\ell^k}=0$ otherwise. The implementation limits this duration so that the expected charging-completion time does not exceed the charging station's upper time-window bound. Because charging durations are deterministic, they do not add variance to the propagated time. The arrival time at the next location is $\widetilde{a}_{v_{\ell+1}^k}=\widetilde{s}_{v_\ell^k} + d_{v_\ell^k} + q_{v_\ell^k}+\widetilde{t}_{v_\ell^k,v_{\ell+1}^k}$.

In the chance-constrained EADARP, lower time-window bounds are enforced through waiting, while upper time-window bounds at pickup and drop-off locations are enforced probabilistically. For a given confidence level $\alpha$, each passenger location must satisfy
\begin{equation}
\Pr\left(
\widetilde{s}_i
\leq
w_i^{\mathrm{end}}
\right)
\geq
\alpha,
\qquad
\forall i\in P\cup D.
\label{eq:cc_eadarp}
\end{equation}
We apply chance constraints only at pickup and drop-off locations, where lateness directly affects passenger service. Time-window constraints at depots and charging stations remain deterministic. Specifically, charging must finish by the charging station's upper bound, and each vehicle must reach its end depot within the planning horizon. We assume $\alpha\in[\alpha_l,\alpha_h]$, where $\alpha_l>1/2$ and $\alpha_h<1$.

If no waiting has occurred at location $i$ or any earlier location on its route, the service-start time at $i$ is a sum of independent normally distributed travel times and deterministic durations. In this case, the chance constraint has the exact deterministic representation
\begin{equation}
\mu_i+z_\alpha\sqrt{v_i}
\leq w_i^{\mathrm{end}},
\qquad \forall i\in P\cup D,
\label{eq:normal_cc_screen}
\end{equation}
where $\mu_i=\mathbb{E}[\widetilde{s}_i]$ and $v_i=\operatorname{Var}(\widetilde{s}_i)$ denote the mean and variance of the service-start time, respectively. Furthermore, $z_\alpha$ denotes the $\alpha$-quantile of the standard normal distribution~\cite{ijcai2022p665,10254112}.

When waiting occurs, the service-start time is no longer normally distributed because all arrival times before the applicable lower bound are replaced by that bound. We therefore calculate its first two moments and use a normal distribution with the same mean and variance to approximate the uncertainty at subsequent locations.

Let $A\sim\mathcal{N}(\mu,v)$, $B=\max(A,c)$, where $A$ denotes the arrival time and $c$ denotes the applicable deterministic lower bound. At a pickup location, $c$ may also include a later service-time bound introduced by the nominal scheduling procedure to satisfy the maximum ride-time constraint. For $v>0$, the mean and the second moment of $B$ are $\mathbb{E}[B]=\mu\Phi(-\eta)+c\Phi(\eta)+\sqrt{v}\,\phi(\eta)$, $\mathbb{E}[B^2]=(\mu^2+v)\Phi(-\eta)+c^2\Phi(\eta)+(\mu+c)\sqrt{v}\,\phi(\eta)$, where $\eta=(c-\mu)/\sqrt{v}$. Here, $\phi$ and $\Phi$ denote the probability density function and cumulative distribution function of the standard normal distribution, respectively~\cite{4403040}. The resulting moments of the service-start time are $\mu_i=\mathbb{E}[B]$, and $v_i=\operatorname{Var}(B)=\mathbb{E}[B^2]-\mathbb{E}[B]^2$.

\section{Insertion-Operator Selection in Bilevel Large Neighborhood Search}
\label{sec:method}

This section presents the baseline BI-LNS framework and the six insertion-operator selection strategies. 

\subsection{BI-LNS framework}
\label{sec:bilns_framework}

To solve both the deterministic and chance-constrained variants of the EADARP, we adapt the bilevel large neighborhood
search framework introduced by Limmer~\cite{LIMMER2023100876}. BI-LNS separates charging and request-routing decisions. The outer level inserts, removes, or modifies charging sessions, while the inner level arranges pickup and drop-off locations around the fixed charging sessions. Each charging session specifies a station, a charging duration, and an upper bound on its completion time.

The algorithm starts with one empty route for each vehicle and generates $C_{\mathrm{init}}$ charging sessions for each route. It then calls the inner search to insert the passenger requests. At each outer iteration, it selects charging session modification, insertion, or removal with probabilities $p_{\mathrm{mod}}$, $p_{\mathrm{ins}}$, and $p_{\mathrm{rem}}$, respectively. If the solution contains no charging sessions, it selects insertion. After changing a route, the algorithm removes all passenger requests from that route and calls the inner search to reinsert them.

We consider unlimited charging-station visits where a station can appear in multiple charging sessions. The outer search prioritizes the number of served requests until it finds a complete solution. 

\subsection{Inner-level search}
\label{sec:inner_lns}

For fixed charging sessions, the inner level applies a destroy-and-repair search. It first orders the requests by increasing lower bound of their pickup time windows and attempts to insert them. Requests that cannot be inserted remain rejected.

Each inner iteration starts from the best retained solution. The destroy step randomly removes $q_{\mathrm{out}}$ served requests. The algorithm combines these requests with the rejected set and selects up to $q_{\mathrm{in}}$ requests for reinsertion. It orders the selected requests by their pickup time-window lower bounds and applies the insertion operator chosen by the corresponding selection strategy.

Before a complete solution is found, the algorithm starts with $Q_{\mathrm{in}}$ and $Q_{\mathrm{out}}$ and adjusts these values according to the number of rejected requests. After all requests are served, it sets $q_{\mathrm{in}}=q_{\mathrm{out}}=Q$. We use the same adjustment in all six variants.

The inner search follows a feasibility-first acceptance rule. While the best solution is incomplete, it accepts a route-feasible candidate if the candidate does not increase the number of rejected requests. It accepts a route-infeasible candidate only when the candidate strictly reduces that number. After finding a complete solution, it accepts only complete candidates that do not increase the objective value. Algorithm~\ref{alg:inner_bilns_selection} summarizes the
inner search.

\begin{algorithm}[t]
\caption{Inner-level BI-LNS with insertion-operator selection}
\label{alg:inner_bilns_selection}
\begin{algorithmic}[1]
\REQUIRE Routes $\mathcal{R}$ with fixed charging sessions, requests $R$,
iteration limit $I$, and selection strategy $m$
\ENSURE Best routes $\mathcal{R}^{*}$ and rejected requests $R^{*}$
\STATE Order $R$ by pickup time-window lower bound
\STATE Select an initial operator using $m$ and insert the requests
\STATE Store the result as $(\mathcal{R}^{*},R^{*})$
\FOR{$h=1,\ldots,I$}
    \STATE Set $q_{\mathrm{out}}$ and $q_{\mathrm{in}}$ from the current
    feasibility state
    \STATE Copy $(\mathcal{R}^{*},R^{*})$ to $(\mathcal{R}',R')$
    \STATE Remove $q_{\mathrm{out}}$ served requests and add them to $R'$
    \STATE Select and order up to $q_{\mathrm{in}}$ requests from $R'$
    \STATE Select $o\in\{\textsc{Greedy},\textsc{Regret}\}$ using $m$
    \STATE Apply $o$ using the feasibility test for the relevant EADARP variant
    \IF{the candidate satisfies the acceptance rule}
        \STATE $(\mathcal{R}^{*},R^{*})\leftarrow(\mathcal{R}',R')$
    \ENDIF
    \STATE Update the information required by $m$
\ENDFOR
\RETURN $(\mathcal{R}^{*},R^{*})$
\end{algorithmic}
\end{algorithm}

For the deterministic EADARP, the algorithm applies the feasibility conditions in Section~\ref{sec:problem_definition}. For the chance-constrained variant, it uses the time-window evaluation in Section~\ref{sec:cc_eadarp}. Therefore, the selection strategy changes only the insertion operator, not the constraints or their evaluation.

\subsection{Insertion operators}
\label{sec:insertion_operators}

We use greedy and regret-based insertion because they provide different repair behaviors. Greedy insertion focuses on the best immediate move, while regret insertion prioritizes requests that may become difficult to insert later.

\subsubsection{Greedy insertion}

The greedy operator processes requests sequentially. For each request, it finds feasible pickup and drop-off positions across the vehicle routes and selects the positions with the smallest increase in the objective value. If it finds no feasible insertion, the request remains rejected. 

\subsubsection{Regret-based insertion}

The regret operator first ranks the rejected requests and then passes the resulting order to greedy insertion~\cite{Hemmelmayr2012}. For each request, it estimates the added travel time for a restricted set of insertion positions. These estimates rank the requests but do not determine feasibility.

Let $\widetilde c_1(r)\leq\cdots\leq\widetilde c_m(r)$ denote the approximate insertion costs for request $r$. With $k=3$, we calculate $\operatorname{Regret}_{k}(r)=\sum_{j=2}^{\min\{k,m\}}\left(\widetilde c_j(r)-\widetilde c_1(r)\right)$. If a request has fewer than two estimated insertion options, we assign it an infinite regret value so that the algorithm considers it first. We rank requests by decreasing regret and randomly shuffle the highest ranked subset to add diversity. Greedy insertion then processes this order and applies the full deterministic or chance-constrained feasibility test.

\subsection{Insertion-operator selection strategies}
\label{sec:operator_selection}

We compare the six insertion-operator selection strategies summarized in Table~\ref{tab:operator_selection}. They differ only in how they select between greedy and regret-$k$ insertion. The \emph{State} column indicates whether the selector uses features of the current search state. The \emph{History} column indicates whether it uses the past performance of the insertion operators. The \emph{Selection} column specifies the operator-selection mechanism.

\begin{table}[t]
\centering
\caption{Insertion-operator selection strategies.}
\label{tab:operator_selection}
\begin{tabular}{llll}
\toprule
Variant & State & History & Selection \\
\midrule
BI-LNS-Greedy & No  & No  & Fixed greedy \\
BI-LNS-Regret & No  & No  & Fixed regret-$k$ \\
BI-LNS-Random & No  & No  & Uniform random \\
BI-LNS-Rule   & Yes & No  & Fixed thresholds \\
BI-LNS-ALNS   & No  & Yes & Adaptive weights \\
BI-LNS-LLM    & Yes & No  & Model preference \\
\bottomrule
\end{tabular}
\end{table}

\paragraph{Fixed and random selection.}
BI-LNS-Greedy always uses greedy insertion, as in the original BI-LNS. BI-LNS-Regret always uses regret-$k$ insertion. BI-LNS-Random selects either operator with equal probability at each decision point, including during the initial construction.

\paragraph{Rule-based selection.}
BI-LNS-Rule uses the unrouted request ratio $r_u$, the mean time-window tightness $\tau$ of unrouted requests, and the recent rejection rate $\rho$. It selects regret-$k$ insertion when $r_u>0.30$, $\tau>0.60$, or $\rho>0.15$. Otherwise, it selects greedy insertion. It uses greedy insertion during the initial construction.

\paragraph{Performance-adaptive selection.}
Inspired by adaptive operator selection in ALNS~\cite{Ropke2006},
BI-LNS-ALNS assigns a weight $w_o$ to each insertion operator and selects operator $o$ with probability $\Pr(o)=w_o/\sum_{o'}w_{o'}$. The operator weights are initialized to $1$ and updated throughout the search. After each repair iteration, it updates the weight of the selected operator as $w_o\leftarrow0.9w_o+0.1\pi$, where $\pi$ is $33$ for an accepted improvement, $1$ for an accepted non-improving result, and $0$ for a rejected result.

\paragraph{LLM-based selection.}
BI-LNS-LLM uses an LLM as an online hyper-heuristic. The model receives a compact summary of the current inner-search state and returns $p_{\mathrm{greedy}}\in[0,1]$, which represents its preference for greedy insertion. For each run and discretized search state, the algorithm deterministically generates $u\in[0,1)$. It selects greedy insertion if
$u<p_{\mathrm{greedy}}$ and regret-$k$ insertion otherwise.

The LLM does not receive complete routes, construct solutions, or evaluate solutions. BI-LNS performs route modification, feasibility checking, objective calculation, charging decisions, and acceptance. The model receives only the aggregated features listed in Table~\ref{tab:llm_features}. The features capture the current feasibility status, recent search progress,
request difficulty, distance from the best-known solution, and charging activity. They provide the LLM with a compact representation of the search state without exposing complete solutions.

\begin{table}[t]
\centering
\caption{Search-state features supplied to the LLM selector.}
\label{tab:llm_features}
\begin{tabular}{p{0.28\textwidth}p{0.65\textwidth}}
\toprule
Feature & Definition \\
\midrule
\texttt{rejected} &
Number of currently rejected requests. \\
\texttt{rejected\_ratio} &
Ratio of rejected requests to all requests. \\
\texttt{delta\_rejected} &
Normalized change in the rejected-request count from the previous iteration. \\
\texttt{delta\_obj} &
Normalized objective improvement from the previous iteration. \\
\texttt{stagnation} &
Normalized number of consecutive iterations without improvement. \\
\texttt{progress} &
Fraction of completed inner-search iterations. \\
\texttt{tw\_tightness} &
Mean normalized time-window tightness of rejected requests. \\
\texttt{unrouted\_hardness} &
Product of \texttt{rejected\_ratio} and \texttt{tw\_tightness}. \\
\texttt{gap\_best\_rejected} &
Normalized difference from the best rejected-request count. \\
\texttt{gap\_best\_obj} &
Normalized difference from the best objective value. \\
\texttt{cs\_count} &
Mean number of charging sessions per vehicle. \\
\bottomrule
\end{tabular}
\end{table}

\begin{figure}[!t]
\centering
\includegraphics[width=0.4\textwidth]{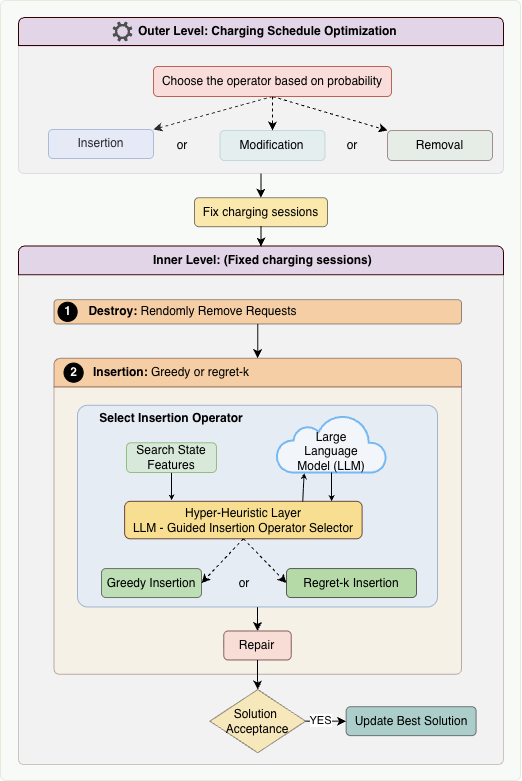}
\caption{Role of the LLM selector in the BI-LNS inner-level repair process.}
\label{fig:architecture}
\end{figure}

We cache the LLM preference for each discretized search state to reduce repeated API calls and reuse the most recently selected operator between model queries. If the model is unavailable or returns an invalid response, BI-LNS-LLM uses the rule-based selector. Figure~\ref{fig:architecture} illustrates the role of the LLM selector.

\section{Experimental Setting}
\label{sec:experimental_setting}

In this section, we present the experimental setup used to evaluate our approach on deterministic  and chance-constrained EADARP instances.

We use two benchmark sets derived from classical DARP instances. The first set contains ten instances introduced in \cite{Cordeau}, and the second contains ten instances introduced in \cite{ropke}. Following \cite{BONGIOVANNI2019436}, we extend these instances by adding battery constraints and three charging stations. We denote an instance by $aK$-$N$, where $K$ is the number of vehicles and $N$ is the number of transportation requests.

For the deterministic experiments, consistent with prior EADARP studies~\cite{LIMMER2023100876,10.1145/3638529.3654057}, we set the minimum final battery ratio to $\gamma=0.7$. We also examine the effect of this parameter on the Ropke instances using $\gamma\in\{0.2,0.4,0.5,0.7,0.8,0.9\}$. Following previous EADARP studies~\cite{LIMMER2023100876, 10.1145/3638529.3654057}, we set the objective weights in Equation~\eqref{eq:objective} to $w_1=0.75$ and $w_2=0.25$.

For the chance-constrained experiments, we use coefficients of variation $\mathrm{CV}\in\{0.1,0.3\}$
and confidence levels $\alpha\in\{0.99,0.9999,0.999999\}$. We consider minimum final battery ratios $\gamma\in\{0.4,0.7\}$.

We implement all methods in Python~3.11 using the BI-LNS framework described in Section~\ref{sec:method}. We set the initial number of charging sessions per vehicle to $C_{\mathrm{init}}=1$. At the outer level, we select charging-session modification, insertion, and removal with probabilities $0.55$, $0.225$, and $0.225$, respectively. Each call to the inner-level search performs $I=300$ iterations, and each run has a time limit of 120 minutes.

Before a complete solution is found, we initially set \(Q_{\mathrm{in}}=2\) and \(Q_{\mathrm{out}}=1\). As explained in Section~\ref{sec:inner_lns}, the algorithm may increase these values according to the number of rejected requests. After a complete deterministic solution is found, we set $Q=\max\{10,\lfloor0.4N\rfloor\}$. The reported chance-constrained experiments use $Q=19$. We apply a noise factor of $\eta=0.01$ to the insertion costs~\cite{LIMMER2023100876} and use $k=3$ for regret insertion~\cite{Hemmelmayr2012}.

These settings follow the main structure of BI-LNS but include changes to the number of inner iterations and the request-removal and reinsertion rules. Therefore, we treat the method as an adapted Python implementation of BI-LNS rather than an exact reproduction of the implementation in \cite{LIMMER2023100876}.

We compare six variants: BI-LNS-Greedy, BI-LNS-Regret, BI-LNS-Random, BI-LNS-Rule, BI-LNS-ALNS, and BI-LNS-LLM. The variants use the same problem representation, outer-level search, feasibility tests, acceptance rule, parameter settings, and stopping criteria. They differ only in the method used to select the inner-level insertion operator.

For BI-LNS-LLM, we use the Gemini API~\footnote{\href{https://ai.google.dev/gemini-api/docs/api-key}{Google AI Studio}} and use the \textit{gemini-2.5-flash-lite} model. We set the temperature to zero and limit the response to 50 tokens.

We run the experiments on single nodes of a high-performance computing cluster equipped with Intel\textsuperscript{\textregistered} Xeon\textsuperscript{\textregistered} Platinum~8360Y processors. Each job is
allocated eight CPU cores and 1~GB of memory. For each experimental configuration, we conduct ten independent runs. We report the feasibility rate and the minimum, mean, and standard deviation of the objective values obtained from feasible runs. Objective statistics should therefore be interpreted together with the feasibility rate.

\section{Experimental Results}
\label{sec:experimental_results}

This section compares the six operator-selection strategies on both the deterministic  and chance-constrained EADARP, focusing on solution quality, feasibility recovery, and search dynamics across benchmark instances.

\begin{landscape}
    \begin{table*}
\centering
\scriptsize
\caption{Final best feasible objective values for the deterministic  EADARP on the Cordeau and Ropke instances with $\gamma=0.7$. For each method, we report the minimum, mean, and standard deviation over the feasible runs among 10 independent runs, together with the feasibility count. Best conditional mean values per instance are highlighted in bold. NA indicates that no feasible run was obtained, and - indicates that the standard deviation is undefined because only one feasible run was obtained.}
\label{tab:results_cordeau_unlimited}
\setlength{\tabcolsep}{2.5pt}
\renewcommand{\arraystretch}{1.05}
\resizebox{\linewidth}{!}{
\begin{tabular}{lrrrrrrrrrrrrrrrrrrrrrrrr}
\toprule
\multirow[c]{2}{*}{Instance} & \multicolumn{4}{c}{BI-LNS-Greedy~\cite{LIMMER2023100876}} & \multicolumn{4}{c}{BI-LNS-Regret} & \multicolumn{4}{c}{BI-LNS-Random} & \multicolumn{4}{c}{BI-LNS-Rule} & \multicolumn{4}{c}{BI-LNS-ALNS} & \multicolumn{4}{c}{BI-LNS-LLM} \\ 
 & Min & Mean & Std & Feas.  & Min & Mean & Std & Feas.  & Min & Mean & Std & Feas.  & Min & Mean & Std & Feas.  & Min & Mean & Std & Feas.  & Min & Mean & Std & Feas.  \\ 
 \midrule
a2-16 & $257.6$ & $258.0$ & $0.4$ & 10/10 & $256.4$ & $257.6$ & $0.8$ & 10/10 & $256.6$ & $\mathbf{257.4}$ & $0.5$ & 10/10 & $256.7$ & $257.9$ & $0.7$ & 10/10 & $256.5$ & $258.1$ & $0.8$ & 10/10 & $257.3$ & $257.9$ & $0.6$ & 10/10 \\
a3-18 & $250.1$ & $\mathbf{250.3}$ & $0.1$ & 10/10 & $250.3$ & $\mathbf{250.3}$ & $0.0$ & 10/10 & $250.1$ & $\mathbf{250.3}$ & $0.1$ & 10/10 & $250.1$ & $\mathbf{250.3}$ & $0.1$ & 10/10 & $250.2$ & $\mathbf{250.3}$ & $0.1$ & 10/10 & $250.1$ & $\mathbf{250.3}$ & $0.1$ & 10/10 \\
a3-24 & $276.3$ & $276.3$ & $0.0$ & 10/10 & $275.2$ & $276.2$ & $0.3$ & 10/10 & $275.2$ & $276.2$ & $0.3$ & 10/10 & $275.2$ & $276.2$ & $0.3$ & 10/10 & $275.2$ & $\mathbf{275.9}$ & $0.5$ & 10/10 & $276.3$ & $276.3$ & $0.0$ & 10/10 \\
a4-16 & $226.8$ & $227.4$ & $0.8$ & 10/10 & $226.8$ & $227.4$ & $0.8$ & 10/10 & $226.8$ & $227.3$ & $0.8$ & 10/10 & $226.8$ & $\mathbf{227.0}$ & $0.5$ & 10/10 & $226.8$ & $\mathbf{227.0}$ & $0.5$ & 10/10 & $226.8$ & $227.5$ & $0.7$ & 10/10 \\
a4-24 & $322.8$ & $324.5$ & $3.8$ & 10/10 & $322.4$ & $\mathbf{322.8}$ & $0.2$ & 10/10 & $322.4$ & $\mathbf{322.8}$ & $0.2$ & 10/10 & $322.4$ & $\mathbf{322.8}$ & $0.2$ & 10/10 & $322.8$ & $322.9$ & $0.1$ & 10/10 & $322.8$ & $323.0$ & $0.1$ & 10/10 \\
a4-32 & $417.9$ & $421.2$ & $2.4$ & 10/10 & $417.9$ & $420.7$ & $1.3$ & 10/10 & $418.3$ & $420.5$ & $1.3$ & 10/10 & $419.6$ & $421.9$ & $1.0$ & 10/10 & $417.9$ & $420.6$ & $1.9$ & 10/10 & $417.9$ & $\mathbf{420.2}$ & $1.8$ & 10/10 \\
a4-40 & $498.1$ & $514.5$ & $11.3$ & 10/10 & $490.5$ & $509.7$ & $10.4$ & 10/10 & $497.8$ & $513.5$ & $12.0$ & 10/10 & $501.9$ & $512.1$ & $10.1$ & 9/10 & $501.8$ & $515.4$ & $9.2$ & 10/10 & $502.9$ & $\mathbf{508.0}$ & $3.9$ & 10/10 \\
a4-48 & $603.2$ & $610.7$ & $6.3$ & 10/10 & $601.1$ & $\mathbf{607.6}$ & $3.9$ & 10/10 & $601.0$ & $610.7$ & $7.1$ & 10/10 & $599.5$ & $609.4$ & $8.5$ & 10/10 & $605.5$ & $609.8$ & $4.1$ & 10/10 & $602.9$ & $609.8$ & $5.4$ & 10/10 \\
a5-40 & $420.7$ & $\mathbf{424.3}$ & $2.3$ & 10/10 & $423.0$ & $424.8$ & $2.2$ & 10/10 & $422.7$ & $426.2$ & $3.7$ & 10/10 & $423.0$ & $425.4$ & $2.4$ & 10/10 & $420.3$ & $424.7$ & $3.0$ & 10/10 & $423.0$ & $427.0$ & $4.7$ & 10/10 \\
a5-50 & $588.6$ & $595.2$ & $4.3$ & 10/10 & $590.6$ & $601.7$ & $8.2$ & 10/10 & $590.5$ & $597.4$ & $6.8$ & 10/10 & $591.4$ & $598.7$ & $8.5$ & 10/10 & $593.2$ & $600.1$ & $4.4$ & 10/10 & $589.6$ & $\mathbf{594.2}$ & $2.8$ & 10/10 \\
\midrule
a5-60 & $770.5$ & $794.8$ & $30.6$ & 8/10 & $769.6$ & $782.3$ & $16.5$ & 8/10 & $753.1$ & $779.7$ & $15.0$ & 8/10 & $756.4$ & $783.5$ & $15.0$ & 9/10 & $755.9$ & $\mathbf{776.6}$ & $17.2$ & 10/10 & $770.3$ & $790.3$ & $21.5$ & 8/10 \\
a6-48 & $528.6$ & $534.6$ & $5.1$ & 10/10 & $527.6$ & $534.6$ & $4.8$ & 10/10 & $527.7$ & $535.2$ & $5.0$ & 10/10 & $529.8$ & $533.1$ & $3.2$ & 10/10 & $530.4$ & $537.0$ & $5.0$ & 10/10 & $527.6$ & $\mathbf{532.9}$ & $2.4$ & 10/10 \\
a6-60 & $738.4$ & $755.2$ & $15.2$ & 10/10 & $741.7$ & $757.3$ & $10.2$ & 10/10 & $743.1$ & $753.3$ & $11.0$ & 10/10 & $739.7$ & $754.2$ & $19.6$ & 10/10 & $743.2$ & $\mathbf{752.3}$ & $7.2$ & 10/10 & $741.6$ & $758.7$ & $15.1$ & 10/10 \\
a6-72 & $874.1$ & $888.7$ & $18.3$ & 5/10 & $875.2$ & $894.5$ & $16.5$ & 8/10 & $877.8$ & $899.5$ & $15.5$ & 10/10 & $859.8$ & $891.5$ & $19.6$ & 10/10 & $859.4$ & $901.9$ & $29.4$ & 9/10 & $867.6$ & $\mathbf{888.6}$ & $18.8$ & 9/10 \\
a7-56 & $647.4$ & $667.5$ & $13.7$ & 10/10 & $651.7$ & $661.2$ & $5.1$ & 10/10 & $646.0$ & $\mathbf{658.5}$ & $6.3$ & 10/10 & $649.0$ & $665.9$ & $13.2$ & 10/10 & $656.9$ & $667.2$ & $11.7$ & 10/10 & $653.8$ & $666.0$ & $8.2$ & 10/10 \\
a7-70 & $837.1$ & $851.2$ & $15.4$ & 10/10 & $835.3$ & $853.4$ & $13.4$ & 10/10 & $827.2$ & $847.5$ & $13.5$ & 10/10 & $833.4$ & $\mathbf{843.5}$ & $11.5$ & 10/10 & $832.3$ & $847.0$ & $10.0$ & 10/10 & $844.5$ & $868.1$ & $37.6$ & 10/10 \\
a7-84 & $970.3$ & $1030.1$ & $31.9$ & 8/10 & $974.3$ & $1005.7$ & $21.9$ & 10/10 & $983.3$ & $1035.0$ & $52.1$ & 7/10 & $968.1$ & $1004.9$ & $31.5$ & 9/10 & $982.5$ & $1008.9$ & $19.5$ & 8/10 & $966.6$ & $\mathbf{998.2}$ & $20.0$ & 7/10 \\
a8-64 & $681.3$ & $697.0$ & $11.6$ & 10/10 & $689.7$ & $695.7$ & $4.7$ & 10/10 & $683.9$ & $697.3$ & $7.6$ & 10/10 & $681.7$ & $696.1$ & $8.4$ & 10/10 & $685.7$ & $693.5$ & $6.1$ & 10/10 & $679.7$ & $\mathbf{691.4}$ & $10.6$ & 10/10 \\
a8-80 & $889.5$ & $917.5$ & $16.3$ & 8/10 & $890.8$ & $\mathbf{915.3}$ & $15.3$ & 9/10 & $891.5$ & $918.9$ & $25.6$ & 9/10 & $897.2$ & $931.8$ & $27.6$ & 9/10 & $880.5$ & $920.6$ & $22.7$ & 9/10 & $911.5$ & $924.7$ & $12.1$ & 8/10 \\
a8-96 & \text{NA} & \text{NA} & \text{NA} & 0/10 & $1229.1$ & $1261.3$ & $45.1$ & 3/10 & $1211.9$ & $1211.9$ & - & 1/10 & $1252.9$ & $1253.0$ & $0.2$ & 2/10 & $1278.0$ & $1278.0$ & - & 1/10 & $1203.1$ & $\mathbf{1203.1}$ & - & 1/10 \\
\bottomrule
\end{tabular}
}
\end{table*}
\end{landscape}

\subsection{Results for the Deterministic  EADARP}
\label{subsec:det_results}

We first evaluate the performance of different inner-level insertion-operator selection methods within BI-LNS on the deterministic  EADARP. 

\begin{figure*}[t]
\centering
\includegraphics[width=\textwidth]
{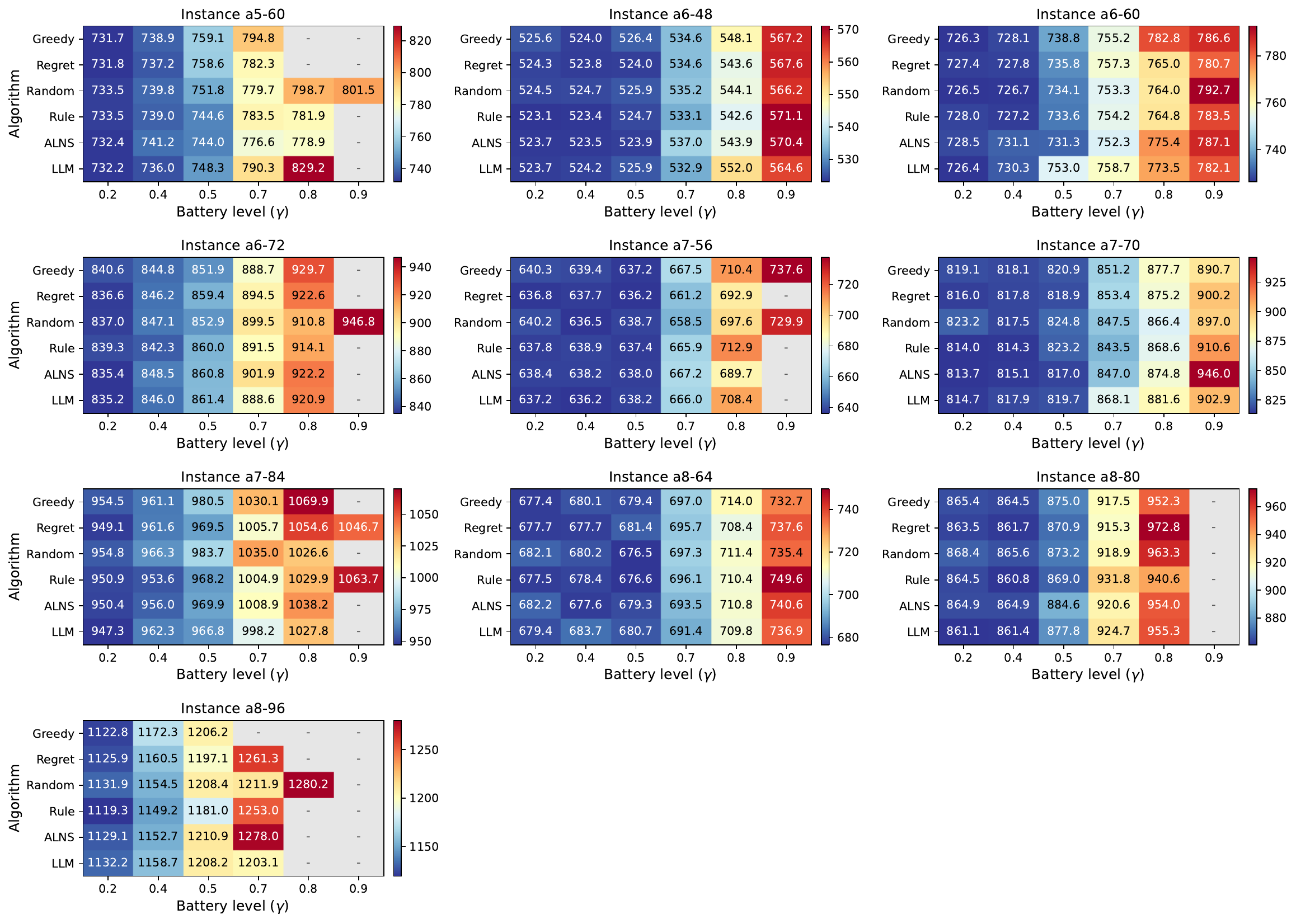}
\caption{Mean objective values over feasible runs for the Ropke instances under different minimum final battery ratios $\gamma\in\{0.2,0.4,0.5,0.7,0.8,0.9\}$. Colours are scaled independently within each instance. Rows represent BI-LNS
variants and columns represent battery levels. Lower values are better, and - indicates that no feasible run was obtained.}
\label{fig:objective_heatmaps}
\end{figure*}

\begin{figure}[t]
\centering
\includegraphics[width=0.45\textwidth]{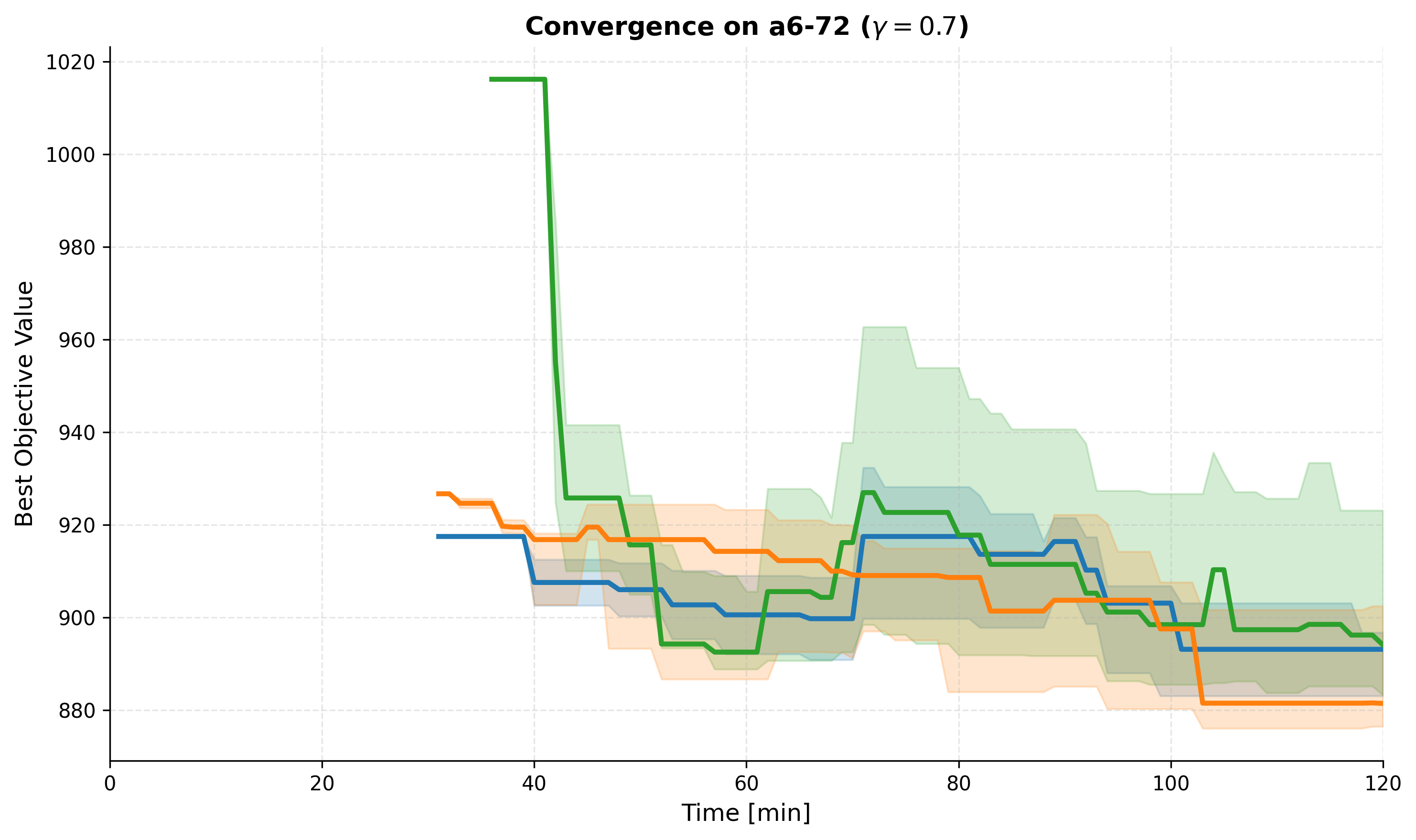}
\includegraphics[width=0.45\textwidth]{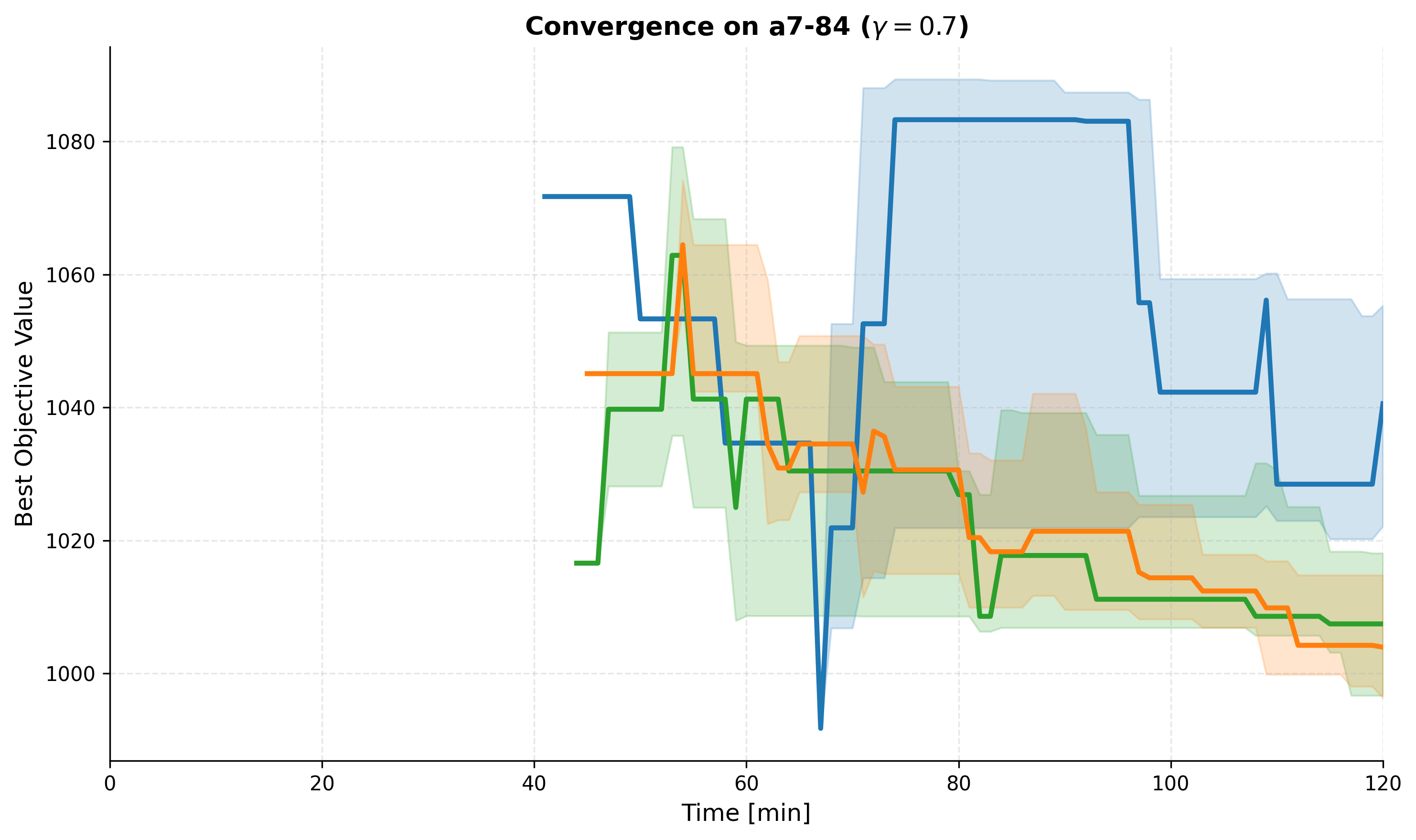}
\includegraphics[width=0.4\textwidth]{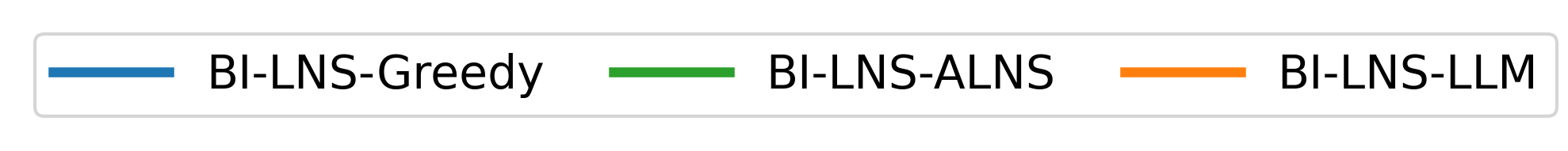}
\caption{Convergence behavior over time on representative hard instances (a6-72 and a7-84), comparing BI-LNS-Greedy, BI-LNS-ALNS, and BI-LNS-LLM. The plots show median best feasible objective values over 10 runs with IQR bands. Only feasible solutions are considered; curves start at the first feasible solution.
}
\label{fig:convergence}
\end{figure}

Table~\ref{tab:results_cordeau_unlimited} presents results for both Cordeau and Ropke instances with $\gamma=0.7$. For each instance and method, we report the minimum, mean, and standard deviation of the best feasible objective values over 10 recorded runs, together with the feasibility rate~(feasible/total runs). Best mean values per instance are highlighted in bold. Entries marked as \textit{NA} indicate that no feasible solution was found within the time limit. When feasibility is not achieved in all runs, objective statistics are computed over feasible runs only and should be interpreted jointly with the feasibility rate.

Across the small and medium-sized instances, the conditional mean differences among the strategies are generally small. On the larger instances, the method with the lowest conditional mean varies by instance. BI-LNS-ALNS obtains the lowest mean on a5-60 and a6-60, BI-LNS-Random on a7-56, BI-LNS-Rule on a7-70, and BI-LNS-Regret on a8-80. BI-LNS-LLM obtains the lowest mean on a6-48, a6-72, a7-84, a8-64, and a8-96. However, a lower conditional mean does not necessarily indicate better overall performance because the objective statistics exclude infeasible runs and the feasibility rates differ across methods.

The feasibility rates further illustrate this trade-off. On a6-72, BI-LNS-Random and BI-LNS-Rule obtain 10/10 feasible runs, whereas BI-LNS-LLM obtains 9/10 with a marginally lower conditional mean. On a7-84, BI-LNS-Regret obtains 10/10 feasible runs, whereas BI-LNS-LLM obtains 7/10 with the lowest conditional mean. Thus, the results suggest that operator selection affects post-feasibility solution quality on some difficult instances, but they do not show that BI-LNS-LLM consistently improves feasibility recovery.
Instance a8-96 is particularly difficult. BI-LNS-Greedy finds no feasible solution, while the other methods obtain only one to three feasible runs. Therefore, the objective values for this instance provide limited comparative evidence. Overall, no selection method performs best across all instances.

\subsubsection{Impact of Minimum Final Battery Ratio \texorpdfstring{$\gamma$}{gamma}}

We analyze the effect of the minimum final battery ratio $\gamma$ on the Ropke instances. Figure~\ref{fig:objective_heatmaps} shows the conditional mean objective value for each BI-LNS variant and battery level. Lower values are better, and blank cells indicate that the corresponding method obtained no
feasible run.

For $\gamma\leq0.5$, most methods obtain similar objective values within each instance. This indicates that the insertion-operator selection strategy has a limited effect when the battery requirement is relatively weak. From $\gamma=0.7$, objective values generally increase because solutions must preserve a higher battery level at the end of each route. Meeting this requirement restricts request allocation, route construction, and charging decisions.

The effect of increasing $\gamma$ depends strongly on the instance. All six methods obtain feasible solutions at $\gamma=0.9$ on a6-48, a6-60, a7-70, and a8-64. In contrast, only BI-LNS-Random obtains a feasible run at $\gamma=0.9$ on a5-60 and a6-72, while only BI-LNS-Greedy and BI-LNS-Random do so on a7-56. On a7-84, only BI-LNS-Regret and BI-LNS-Rule obtain feasible runs at this battery level. No method obtains a feasible run at $\gamma=0.9$ on a8-80 or a8-96. At $\gamma=0.8$ on a8-96, only BI-LNS-Random obtains a feasible run.

These results show that stricter battery requirements increase both objective values and the difficulty of feasibility recovery. However, the method that remains feasible under the strictest setting varies across instances. Therefore, there is no single selection strategy that performs best across all instances and battery settings. We provide detailed objective values and feasibility rates in the Appendix~\ref{app_battery}.

\subsubsection{Search Dynamics and Convergence Behavior}

To examine search behavior, we compare BI-LNS-Greedy, BI-LNS-ALNS, and BI-LNS-LLM on the difficult instances a6-72 and a7-84. Figure~\ref{fig:convergence} shows the median best feasible objective and interquartile range over time. At each time point, we include only runs that have found a feasible solution.

On a6-72, BI-LNS-Greedy obtains a lower median during the middle stage of the search, while BI-LNS-LLM continues to improve during the final stage and obtains the lowest median at the time limit. BI-LNS-ALNS also improves, but its late-stage interquartile range is wider. On a7-84, the three methods show substantial variation after recovering feasibility. After approximately 80 minutes, BI-LNS-LLM and BI-LNS-ALNS achieve lower median objective values than BI-LNS-Greedy. BI-LNS-LLM obtains the lowest median at the end of the search, while BI-LNS-Greedy shows the widest late-stage interquartile range.

\subsection{Results for the Chance-Constrained EADARP}
\label{subsec:cc_results}

\begin{table*}
\centering
\caption{Final best feasible objective values for the chance-constrained EADARP with $CV=0.1$ and $\gamma\in\{0.4,0.7\}$. For each method and confidence level $\alpha$, we report the mean and standard deviation over the feasible runs among 10 independent runs, together with the feasibility count. Best conditional mean values per $(\text{instance},\gamma,\alpha)$ combination are highlighted in bold.}
\label{tab:cc_results_unlimited_cv0.1}
\resizebox{\textwidth}{!}{%
\begin{tabular}{ll*{9}{r}!{\vrule width 1pt}*{9}{r}}
\toprule
\multirow[c]{3}{*}{Instance} & \multirow[c]{3}{*}{$\alpha$} & \multicolumn{9}{c!{\vrule width 1pt}}{$\gamma=0.4$} & \multicolumn{9}{c}{$\gamma=0.7$} \\
& & \multicolumn{3}{c}{BI-LNS-Greedy~\cite{LIMMER2023100876}} & \multicolumn{3}{c}{BI-LNS-ALNS} & \multicolumn{3}{c!{\vrule width 1pt}}{BI-LNS-LLM} & \multicolumn{3}{c}{BI-LNS-Greedy~\cite{LIMMER2023100876}} & \multicolumn{3}{c}{BI-LNS-ALNS} & \multicolumn{3}{c}{BI-LNS-LLM} \\
 &  & Mean & Std & Feas. & Mean & Std & Feas. & Mean & Std & Feas. & Mean & Std & Feas. & Mean & Std & Feas. & Mean & Std & Feas. \\
\midrule
\multirow[c]{3}{*}{a3-18} & 0.99 & $\mathbf{253.1}$ & $0.0$ & 10/10 & $\mathbf{253.1}$ & $0.0$ & 10/10 & $\mathbf{253.1}$ & $0.0$ & 10/10 & $\mathbf{255.1}$ & $2.2$ & 10/10 & $255.2$ & $1.9$ & 10/10 & $255.3$ & $2.0$ & 10/10 \\
 & 0.9999 & $\mathbf{255.7}$ & $0.0$ & 10/10 & $\mathbf{255.7}$ & $0.0$ & 10/10 & $\mathbf{255.7}$ & $0.0$ & 10/10 & $258.5$ & $2.3$ & 10/10 & $\mathbf{257.8}$ & $1.9$ & 10/10 & $\mathbf{257.8}$ & $1.9$ & 10/10 \\
 & 0.999999 & $\mathbf{255.7}$ & $0.0$ & 10/10 & $\mathbf{255.7}$ & $0.0$ & 10/10 & $\mathbf{255.7}$ & $0.0$ & 10/10 & $\mathbf{257.9}$ & $2.0$ & 10/10 & $258.8$ & $2.6$ & 10/10 & $258.1$ & $1.8$ & 10/10 \\
\midrule
\multirow[c]{3}{*}{a3-24} & 0.99 & $278.9$ & $1.4$ & 10/10 & $279.9$ & $2.2$ & 10/10 & $\mathbf{278.5}$ & $1.2$ & 10/10 & $282.4$ & $3.5$ & 10/10 & $\mathbf{280.3}$ & $2.3$ & 10/10 & $281.3$ & $3.8$ & 10/10 \\
 & 0.9999 & $279.3$ & $1.7$ & 10/10 & $\mathbf{278.2}$ & $1.5$ & 10/10 & $279.5$ & $1.7$ & 10/10 & $283.3$ & $4.7$ & 10/10 & $281.9$ & $4.1$ & 10/10 & $\mathbf{281.1}$ & $3.0$ & 10/10 \\
 & 0.999999 & $279.7$ & $1.6$ & 10/10 & $280.1$ & $1.5$ & 10/10 & $\mathbf{278.9}$ & $1.1$ & 10/10 & $284.2$ & $4.2$ & 10/10 & $282.8$ & $3.7$ & 10/10 & $\mathbf{282.2}$ & $3.5$ & 10/10 \\
\cmidrule{1-20}
\multirow[c]{3}{*}{a4-16} & 0.99 & $228.4$ & $0.0$ & 10/10 & $\mathbf{228.3}$ & $0.5$ & 10/10 & $228.4$ & $0.2$ & 10/10 & $228.4$ & $1.3$ & 10/10 & $228.4$ & $0.0$ & 10/10 & $\mathbf{228.3}$ & $0.3$ & 10/10 \\
 & 0.9999 & $\mathbf{232.5}$ & $0.5$ & 10/10 & $\mathbf{232.5}$ & $0.5$ & 10/10 & $232.6$ & $0.3$ & 10/10 & $\mathbf{232.4}$ & $0.5$ & 10/10 & $232.5$ & $0.5$ & 10/10 & $\mathbf{232.4}$ & $0.7$ & 10/10 \\
 & 0.999999 & $\mathbf{232.5}$ & $0.5$ & 10/10 & $\mathbf{232.5}$ & $0.5$ & 10/10 & $232.7$ & $0.0$ & 10/10 & $232.7$ & $0.0$ & 10/10 & $\mathbf{232.2}$ & $0.7$ & 10/10 & $232.3$ & $0.7$ & 10/10 \\
\cmidrule{1-20}
\multirow[c]{3}{*}{a4-24} & 0.99 & $\mathbf{324.5}$ & $0.3$ & 10/10 & $\mathbf{324.5}$ & $0.3$ & 10/10 & $\mathbf{324.5}$ & $0.3$ & 10/10 & $341.5$ & $6.2$ & 10/10 & $342.1$ & $6.6$ & 10/10 & $\mathbf{340.0}$ & $4.5$ & 10/10 \\
 & 0.9999 & $\mathbf{342.0}$ & $0.0$ & 10/10 & $\mathbf{342.0}$ & $0.0$ & 10/10 & $\mathbf{342.0}$ & $0.0$ & 10/10 & $359.2$ & $7.3$ & 4/10 & $358.8$ & $4.6$ & 9/10 & $\mathbf{357.4}$ & $6.3$ & 8/10 \\
 & 0.999999 & $\mathbf{342.0}$ & $0.0$ & 10/10 & $\mathbf{342.0}$ & $0.0$ & 10/10 & $\mathbf{342.0}$ & $0.0$ & 10/10 & $363.4$ & $13.7$ & 4/10 & $354.8$ & $4.0$ & 8/10 & $\mathbf{351.3}$ & $2.1$ & 3/10 \\
\cmidrule{1-20}
\multirow[c]{3}{*}{a5-40} & 0.99 & $430.5$ & $2.4$ & 10/10 & $\mathbf{428.1}$ & $1.8$ & 10/10 & $429.5$ & $2.1$ & 10/10 & $439.4$ & $5.5$ & 10/10 & $\mathbf{438.0}$ & $4.9$ & 10/10 & $438.5$ & $5.7$ & 10/10 \\
 & 0.9999 & $444.2$ & $3.6$ & 10/10 & $444.0$ & $4.5$ & 10/10 & $\mathbf{443.0}$ & $3.1$ & 10/10 & $458.5$ & $6.5$ & 10/10 & $\mathbf{455.6}$ & $3.2$ & 10/10 & $458.4$ & $8.0$ & 10/10 \\
 & 0.999999 & $443.4$ & $4.3$ & 10/10 & $443.6$ & $4.2$ & 10/10 & $\mathbf{443.2}$ & $2.7$ & 10/10 & $460.2$ & $6.2$ & 10/10 & $461.2$ & $7.1$ & 10/10 & $\mathbf{457.0}$ & $7.8$ & 10/10 \\
\cmidrule{1-20}
\multirow[c]{3}{*}{a5-50} & 0.99 & $588.8$ & $4.8$ & 10/10 & $589.9$ & $4.8$ & 10/10 & $\mathbf{586.1}$ & $4.2$ & 10/10 & $628.6$ & $17.4$ & 8/10 & $627.1$ & $18.4$ & 9/10 & $\mathbf{623.4}$ & $17.2$ & 10/10 \\
 & 0.9999 & $594.8$ & $5.8$ & 10/10 & $\mathbf{590.3}$ & $5.1$ & 10/10 & $591.9$ & $5.9$ & 10/10 & $637.0$ & $12.5$ & 5/10 & $\mathbf{630.0}$ & $13.1$ & 9/10 & $630.5$ & $9.8$ & 10/10 \\
 & 0.999999 & $598.6$ & $5.3$ & 10/10 & $\mathbf{592.5}$ & $4.7$ & 10/10 & $596.4$ & $3.6$ & 10/10 & $649.7$ & $18.8$ & 7/10 & $\mathbf{648.6}$ & $13.1$ & 10/10 & $650.1$ & $23.2$ & 8/10 \\
 \cmidrule{1-20}
\multirow[c]{3}{*}{a6-48} & 0.99 & $537.4$ & $5.7$ & 10/10 & $538.1$ & $5.8$ & 10/10 & $\mathbf{535.1}$ & $3.4$ & 10/10 & $\mathbf{555.5}$ & $9.3$ & 10/10 & $556.4$ & $10.4$ & 10/10 & $556.9$ & $10.1$ & 10/10 \\
 & 0.9999 & $543.3$ & $6.2$ & 10/10 & $541.7$ & $5.9$ & 10/10 & $\mathbf{539.9}$ & $6.0$ & 10/10 & $562.1$ & $10.9$ & 10/10 & $\mathbf{558.4}$ & $5.1$ & 10/10 & $560.5$ & $8.3$ & 10/10 \\
 & 0.999999 & $551.5$ & $5.8$ & 10/10 & $551.1$ & $7.3$ & 10/10 & $\mathbf{547.5}$ & $5.4$ & 10/10 & $572.4$ & $9.1$ & 10/10 & $572.0$ & $12.0$ & 10/10 & $\mathbf{565.0}$ & $9.2$ & 10/10 \\
\cmidrule{1-20}
\multirow[c]{3}{*}{a6-60} & 0.99 & $747.2$ & $6.9$ & 10/10 & $\mathbf{743.8}$ & $6.2$ & 10/10 & $743.9$ & $5.7$ & 10/10 & $797.9$ & $20.6$ & 10/10 & $799.1$ & $23.9$ & 10/10 & $\mathbf{785.7}$ & $12.8$ & 10/10 \\
 & 0.9999 & $793.9$ & $5.1$ & 6/10 & $\mathbf{784.9}$ & $3.1$ & 10/10 & $787.6$ & $9.3$ & 10/10 & $\mathbf{830.8}$ & $20.5$ & 7/10 & $845.3$ & $34.0$ & 10/10 & $848.4$ & $19.6$ & 9/10 \\
 & 0.999999 & $815.4$ & $19.7$ & 6/10 & $\mathbf{799.0}$ & $6.5$ & 9/10 & $803.1$ & $19.8$ & 8/10 & $\mathbf{861.0}$ & $12.4$ & 4/10 & $878.4$ & $16.6$ & 4/10 & $888.9$ & $28.0$ & 3/10 \\
\cmidrule{1-20}
\multirow[c]{3}{*}{a7-56} & 0.99 & $646.8$ & $5.7$ & 10/10 & $\mathbf{644.9}$ & $4.9$ & 10/10 & $646.2$ & $3.9$ & 10/10 & $686.7$ & $18.2$ & 10/10 & $682.8$ & $10.5$ & 10/10 & $\mathbf{681.4}$ & $12.2$ & 10/10 \\
 & 0.9999 & $662.2$ & $3.9$ & 10/10 & $\mathbf{661.3}$ & $4.4$ & 10/10 & $661.7$ & $3.4$ & 10/10 & $711.5$ & $18.1$ & 8/10 & $706.1$ & $15.2$ & 10/10 & $\mathbf{703.2}$ & $14.4$ & 9/10 \\
 & 0.999999 & $664.5$ & $5.4$ & 10/10 & $\mathbf{660.6}$ & $5.5$ & 10/10 & $661.9$ & $4.8$ & 10/10 & $715.0$ & $24.0$ & 8/10 & $\mathbf{703.9}$ & $17.2$ & 9/10 & $705.0$ & $26.6$ & 10/10 \\
\cmidrule{1-20}
\multirow[c]{3}{*}{a7-70} & 0.99 & $838.6$ & $6.1$ & 9/10 & $835.4$ & $6.8$ & 10/10 & $\mathbf{834.2}$ & $5.5$ & 10/10 & $893.2$ & $24.5$ & 10/10 & $\mathbf{879.9}$ & $21.9$ & 10/10 & $888.3$ & $14.5$ & 10/10 \\
 & 0.9999 & $\mathbf{843.2}$ & $10.0$ & 10/10 & $846.4$ & $7.4$ & 10/10 & $843.8$ & $6.6$ & 10/10 & $899.0$ & $15.6$ & 8/10 & $\mathbf{891.3}$ & $15.1$ & 10/10 & $901.4$ & $24.7$ & 8/10 \\
 & 0.999999 & $876.1$ & $5.6$ & 10/10 & $\mathbf{869.4}$ & $9.9$ & 10/10 & $870.8$ & $6.9$ & 10/10 & $940.4$ & $14.3$ & 7/10 & $936.9$ & $25.1$ & 7/10 & $\mathbf{934.0}$ & $21.5$ & 8/10 \\
\cmidrule{1-20}
\multirow[c]{3}{*}{a8-64} & 0.99 & $698.7$ & $4.5$ & 10/10 & $\mathbf{698.4}$ & $5.7$ & 10/10 & $699.3$ & $9.8$ & 10/10 & $719.5$ & $7.1$ & 10/10 & $719.6$ & $14.8$ & 10/10 & $\mathbf{715.9}$ & $8.5$ & 10/10 \\
 & 0.9999 & $\mathbf{701.5}$ & $6.5$ & 10/10 & $704.4$ & $6.5$ & 10/10 & $702.1$ & $5.8$ & 10/10 & $724.7$ & $11.0$ & 10/10 & $732.8$ & $12.1$ & 10/10 & $\mathbf{722.5}$ & $7.7$ & 10/10 \\
 & 0.999999 & $707.7$ & $6.8$ & 10/10 & $703.9$ & $6.3$ & 10/10 & $\mathbf{703.3}$ & $7.9$ & 10/10 & $729.6$ & $11.3$ & 10/10 & $\mathbf{727.7}$ & $5.5$ & 10/10 & $732.7$ & $12.8$ & 10/10 \\
\bottomrule
\end{tabular}
}
\end{table*}

Next, we evaluate the performance of different inner-level operator selection strategies for BI-LNS on the chance-constrained EADARP. Table~\ref{tab:cc_results_unlimited_cv0.1} reports results on a subset of Cordeau and Ropke benchmark instances for varying confidence levels $\alpha$ under $CV = 0.1$, with results shown for $\gamma \in \{0.4, 0.7\}$. The selected instances vary from small to large problem sizes. We compare the performance of BI-LNS-Greedy~\cite{LIMMER2023100876}, BI-LNS-ALNS and BI-LNS-LLM methods. 

For $\gamma = 0.4$, all methods achieve high feasibility across small- and
medium-sized instances for all considered confidence levels. In this case, there are no clear differences in objective values between operator selection
strategies, indicating that operator choice has a relatively minor impact when battery constraints are loose and feasibility is easy to maintain.

As $\gamma$ increases to $0.7$, the problem becomes more constrained and
feasibility decreases, particularly for larger instances and higher confidence
levels. Under these conditions, objective values generally increase.
BI-LNS-Greedy shows lower feasibility in several difficult configurations,
while BI-LNS-ALNS and BI-LNS-LLM recover feasibility more often in some of
these cases. Across the considered instances, both BI-LNS-ALNS and BI-LNS-LLM obtain the best mean objective values in several configurations, while BI-LNS-Greedy performs best in a smaller number of cases. However, the method with the best mean does not always achieve the highest feasibility rate. BI-LNS-ALNS provides higher feasibility in several strict settings, whereas BI-LNS-LLM obtains better conditional means in others. Thus, no method consistently performs best in terms of solution quality, feasibility, and variability.

As the confidence level $\alpha$ increases, feasibility decreases due to stricter chance constraints. However, objective values do not vary monotonically with $\alpha$ because the objective is evaluated using mean travel times, while higher confidence levels only restrict the feasible set without directly penalizing the objective. Overall, as $\alpha$, $\gamma$, and problem size increase, the chance-constrained EADARP becomes more challenging, and the impact of operator selection becomes more pronounced.

Results for higher uncertainty levels ($CV = 0.3$) show near-zero feasibility at strict confidence levels across all methods within the given time limit and are presented in the Appendix~\ref{app_cc}.

\section{Conclusions}
\label{sec:conclusion}

In this paper, we investigated the deterministic and chance-constrained variants
of the electric autonomous dial-a-ride problem. In the chance-constrained
setting, we modeled travel times as independent normal random variables and
enforced upper time-window constraints using deterministic approximations of
chance constraints. We compared fixed greedy, fixed regret-$k$, random,
rule-based, performance-adaptive, and LLM-based inner level insertion operator selection within a common BI-LNS framework.

In the deterministic problem, the computational results show that operator selection has little effect on
small instances, where all methods recover feasibility. Its effect is larger on more constrained instances. No strategy consistently achieves both
the highest feasibility rate and the lowest conditional mean objective.
BI-LNS-LLM obtains the lowest conditional mean on several instances, while the other strategies perform best on others and sometimes
recover feasibility more frequently.

In the chance-constrained experiments, feasibility generally decreases as the final battery requirement, confidence level, and travel-time variability
increase. BI-LNS-ALNS and BI-LNS-LLM obtain competitive conditional objective
values in several configurations, but neither method consistently dominates
in solution quality or feasibility. Overall, LLM-based selection provides a state-aware alternative for choosing insertion operators, but it does not consistently outperform simpler selection strategies. Future work will explore enhanced state representations, alternative prompting strategies, and extensions to other stochastic and dynamic vehicle routing problems.

\bibliographystyle{unsrtnat}
\bibliography{library} 
\appendix

\section{Results for Impact of Different Batteries Levels}
\label{app_battery}
Tables~\ref{tab:results_summary_unlimited_0.2}-~\ref{tab:results_summary_unlimited_0.9}, show the results corresponding to the Figure 2 in the main paper for the different minimum final battery ratios from $\gamma=0.2$ to $0.9$. We report the minimum, mean, and standard deviation
of the best feasible objective values over $10$ independent runs, together with
the feasibility rate~(feasible/total runs). Best mean values per instance are
highlighted in bold. Entries marked as NA indicate that no feasible solution was
found within the time limit. 

\begin{landscape}

\begin{table*}
\centering
\scriptsize
\setlength{\tabcolsep}{3pt}
\caption{Final best feasible objective values for Ropke instances for unlimited CS visits ($\gamma=0.2$).}
\label{tab:results_summary_unlimited_0.2}
\resizebox{\linewidth}{!}{%
\begin{tabular}{lrrrrrrrrrrrrrrrrrrrrrrrr}
\toprule
Instance & \multicolumn{4}{c}{BI-LNS-Greedy} & \multicolumn{4}{c}{BI-LNS-Regret} & \multicolumn{4}{c}{BI-LNS-Random} & \multicolumn{4}{c}{BI-LNS-Rule} & \multicolumn{4}{c}{BI-LNS-ALNS} & \multicolumn{4}{c}{BI-LNS-LLM} \\
 & Min & Mean & Std & Feas.  & Min & Mean & Std & Feas.  & Min & Mean & Std & Feas.  & Min & Mean & Std & Feas.  & Min & Mean & Std & Feas.  & Min & Mean & Std & Feas.  \\
\midrule
a5-60 & $723.9$ & $\mathbf{731.7}$ & $4.3$ & 10/10 & $726.0$ & $731.8$ & $3.2$ & 10/10 & $724.7$ & $733.5$ & $8.9$ & 10/10 & $727.5$ & $733.5$ & $4.1$ & 10/10 & $724.9$ & $732.4$ & $4.5$ & 10/10 & $726.3$ & $732.2$ & $6.2$ & 10/10 \\
a6-48 & $522.9$ & $525.6$ & $1.9$ & 10/10 & $522.0$ & $524.3$ & $3.4$ & 10/10 & $521.8$ & $524.5$ & $2.3$ & 10/10 & $522.0$ & $\mathbf{523.1}$ & $1.5$ & 10/10 & $522.0$ & $523.7$ & $2.1$ & 10/10 & $521.8$ & $523.7$ & $2.1$ & 10/10 \\
a6-60 & $720.0$ & $\mathbf{726.3}$ & $4.4$ & 10/10 & $723.9$ & $727.4$ & $2.5$ & 10/10 & $721.1$ & $726.5$ & $4.0$ & 10/10 & $724.3$ & $728.0$ & $3.9$ & 10/10 & $722.8$ & $728.5$ & $5.0$ & 10/10 & $722.1$ & $726.4$ & $4.9$ & 10/10 \\
a6-72 & $829.5$ & $840.6$ & $9.5$ & 10/10 & $828.1$ & $836.6$ & $5.8$ & 10/10 & $828.8$ & $837.0$ & $6.0$ & 10/10 & $824.7$ & $839.3$ & $10.9$ & 10/10 & $830.0$ & $835.4$ & $3.4$ & 10/10 & $827.0$ & $\mathbf{835.2}$ & $6.7$ & 10/10 \\
a7-56 & $635.4$ & $640.3$ & $4.1$ & 10/10 & $634.2$ & $\mathbf{636.8}$ & $2.6$ & 10/10 & $632.1$ & $640.2$ & $4.5$ & 10/10 & $633.1$ & $637.8$ & $2.6$ & 10/10 & $635.9$ & $638.4$ & $2.8$ & 10/10 & $634.4$ & $637.2$ & $2.0$ & 10/10 \\
a7-70 & $809.3$ & $819.1$ & $8.2$ & 10/10 & $806.1$ & $816.0$ & $7.4$ & 10/10 & $809.5$ & $823.2$ & $9.3$ & 10/10 & $805.6$ & $814.0$ & $6.2$ & 10/10 & $800.6$ & $\mathbf{813.7}$ & $7.2$ & 10/10 & $803.9$ & $814.7$ & $11.0$ & 10/10 \\
a7-84 & $936.8$ & $954.5$ & $10.2$ & 10/10 & $933.1$ & $949.1$ & $13.3$ & 10/10 & $938.2$ & $954.8$ & $13.2$ & 10/10 & $937.3$ & $950.9$ & $8.5$ & 10/10 & $934.7$ & $950.4$ & $11.3$ & 10/10 & $938.1$ & $\mathbf{947.3}$ & $8.8$ & 10/10 \\
a8-64 & $669.5$ & $\mathbf{677.4}$ & $6.2$ & 10/10 & $667.4$ & $677.7$ & $4.7$ & 10/10 & $675.7$ & $682.1$ & $4.7$ & 10/10 & $671.2$ & $677.5$ & $4.9$ & 10/10 & $676.9$ & $682.2$ & $6.1$ & 10/10 & $668.4$ & $679.4$ & $6.7$ & 10/10 \\
a8-80 & $856.6$ & $865.4$ & $7.5$ & 10/10 & $857.6$ & $863.5$ & $6.7$ & 10/10 & $856.7$ & $868.4$ & $12.6$ & 10/10 & $855.6$ & $864.5$ & $7.9$ & 10/10 & $855.0$ & $864.9$ & $7.5$ & 10/10 & $850.4$ & $\mathbf{861.1}$ & $7.2$ & 10/10 \\
a8-96 & $1096.4$ & $1122.8$ & $13.9$ & 10/10 & $1107.3$ & $1125.9$ & $17.0$ & 10/10 & $1096.3$ & $1131.9$ & $17.4$ & 10/10 & $1095.9$ & $\mathbf{1119.3}$ & $13.8$ & 10/10 & $1102.9$ & $1129.1$ & $20.5$ & 10/10 & $1106.0$ & $1132.2$ & $12.0$ & 10/10 \\
\bottomrule
\end{tabular}
}
\end{table*}
\begin{table*}
\centering
\scriptsize
\setlength{\tabcolsep}{3pt}
\caption{Final best feasible objective values for Ropke instances for unlimited CS visits ($\gamma=0.4$).}
\label{tab:results_summary_unlimited_0.4}
\resizebox{\linewidth}{!}{%
\begin{tabular}{lrrrrrrrrrrrrrrrrrrrrrrrr}
\toprule
Instance & \multicolumn{4}{c}{BI-LNS-Greedy} & \multicolumn{4}{c}{BI-LNS-Regret} & \multicolumn{4}{c}{BI-LNS-Random} & \multicolumn{4}{c}{BI-LNS-Rule} & \multicolumn{4}{c}{BI-LNS-ALNS} & \multicolumn{4}{c}{BI-LNS-LLM} \\
 & Min & Mean & Std & Feas.  & Min & Mean & Std & Feas.  & Min & Mean & Std & Feas.  & Min & Mean & Std & Feas.  & Min & Mean & Std & Feas.  & Min & Mean & Std & Feas.  \\
\midrule
a5-60 & $731.1$ & $738.9$ & $7.6$ & 10/10 & $728.6$ & $737.2$ & $5.4$ & 10/10 & $728.6$ & $739.8$ & $9.5$ & 10/10 & $731.3$ & $739.0$ & $4.4$ & 10/10 & $726.8$ & $741.2$ & $8.1$ & 10/10 & $731.4$ & $\mathbf{736.0}$ & $3.2$ & 10/10 \\
a6-48 & $521.8$ & $524.0$ & $1.6$ & 10/10 & $522.0$ & $523.8$ & $1.5$ & 10/10 & $522.9$ & $524.7$ & $0.9$ & 10/10 & $521.8$ & $\mathbf{523.4}$ & $1.5$ & 10/10 & $521.8$ & $523.5$ & $1.0$ & 10/10 & $521.8$ & $524.2$ & $1.6$ & 10/10 \\
a6-60 & $722.6$ & $728.1$ & $4.3$ & 10/10 & $721.4$ & $727.8$ & $3.7$ & 10/10 & $720.8$ & $\mathbf{726.7}$ & $4.0$ & 10/10 & $722.6$ & $727.2$ & $2.7$ & 10/10 & $725.9$ & $731.1$ & $3.6$ & 10/10 & $722.2$ & $730.3$ & $4.9$ & 10/10 \\
a6-72 & $834.6$ & $844.8$ & $8.1$ & 10/10 & $835.2$ & $846.2$ & $7.5$ & 10/10 & $837.0$ & $847.1$ & $7.6$ & 10/10 & $830.7$ & $\mathbf{842.3}$ & $7.3$ & 10/10 & $838.1$ & $848.5$ & $7.7$ & 10/10 & $831.1$ & $846.0$ & $11.4$ & 10/10 \\
a7-56 & $633.5$ & $639.4$ & $3.6$ & 10/10 & $633.5$ & $637.7$ & $3.5$ & 10/10 & $634.2$ & $636.5$ & $2.0$ & 10/10 & $632.5$ & $638.9$ & $4.8$ & 10/10 & $634.4$ & $638.2$ & $4.2$ & 10/10 & $631.8$ & $\mathbf{636.2}$ & $2.9$ & 10/10 \\
a7-70 & $809.3$ & $818.1$ & $5.1$ & 10/10 & $806.2$ & $817.8$ & $8.5$ & 10/10 & $808.8$ & $817.5$ & $6.9$ & 10/10 & $805.8$ & $\mathbf{814.3}$ & $5.8$ & 10/10 & $804.6$ & $815.1$ & $7.3$ & 10/10 & $806.8$ & $817.9$ & $8.1$ & 10/10 \\
a7-84 & $939.1$ & $961.1$ & $10.3$ & 10/10 & $951.7$ & $961.6$ & $9.6$ & 10/10 & $948.7$ & $966.3$ & $11.3$ & 10/10 & $944.1$ & $\mathbf{953.6}$ & $7.3$ & 10/10 & $940.4$ & $956.0$ & $10.3$ & 10/10 & $951.1$ & $962.3$ & $7.6$ & 10/10 \\
a8-64 & $671.0$ & $680.1$ & $7.8$ & 10/10 & $671.6$ & $677.7$ & $5.1$ & 10/10 & $675.0$ & $680.2$ & $3.3$ & 10/10 & $671.4$ & $678.4$ & $5.3$ & 10/10 & $669.0$ & $\mathbf{677.6}$ & $5.3$ & 10/10 & $676.7$ & $683.7$ & $3.8$ & 10/10 \\
a8-80 & $848.1$ & $864.5$ & $8.0$ & 10/10 & $855.3$ & $861.7$ & $4.4$ & 10/10 & $852.4$ & $865.6$ & $6.3$ & 10/10 & $853.8$ & $\mathbf{860.8}$ & $4.8$ & 10/10 & $852.1$ & $864.9$ & $8.8$ & 10/10 & $848.6$ & $861.4$ & $6.7$ & 10/10 \\
a8-96 & $1129.0$ & $1172.3$ & $29.4$ & 10/10 & $1124.9$ & $1160.5$ & $28.5$ & 10/10 & $1134.2$ & $1154.5$ & $13.8$ & 10/10 & $1130.2$ & $\mathbf{1149.2}$ & $10.1$ & 10/10 & $1118.1$ & $1152.7$ & $17.3$ & 10/10 & $1139.7$ & $1158.7$ & $14.5$ & 10/10 \\
\bottomrule
\end{tabular}
}
\end{table*}
\begin{table*}
\centering
\scriptsize
\setlength{\tabcolsep}{3pt}
\caption{Final best feasible objective values for Ropke instances for unlimited CS visits ($\gamma=0.5$).}
\label{tab:results_summary_unlimited_0.5}
\resizebox{\linewidth}{!}{%
\begin{tabular}{lrrrrrrrrrrrrrrrrrrrrrrrr}
\toprule
Instance & \multicolumn{4}{c}{BI-LNS-Greedy} & \multicolumn{4}{c}{BI-LNS-Regret} & \multicolumn{4}{c}{BI-LNS-Random} & \multicolumn{4}{c}{BI-LNS-Rule} & \multicolumn{4}{c}{BI-LNS-ALNS} & \multicolumn{4}{c}{BI-LNS-LLM} \\
 & Min & Mean & Std & Feas.  & Min & Mean & Std & Feas.  & Min & Mean & Std & Feas.  & Min & Mean & Std & Feas.  & Min & Mean & Std & Feas.  & Min & Mean & Std & Feas.  \\
\midrule
a5-60 & $741.8$ & $759.1$ & $10.9$ & 10/10 & $733.0$ & $758.6$ & $20.8$ & 10/10 & $731.4$ & $751.8$ & $13.5$ & 10/10 & $735.4$ & $744.6$ & $6.0$ & 10/10 & $732.7$ & $\mathbf{744.0}$ & $6.7$ & 10/10 & $738.0$ & $748.3$ & $6.5$ & 10/10 \\
a6-48 & $523.8$ & $526.4$ & $2.3$ & 10/10 & $521.8$ & $524.0$ & $1.7$ & 10/10 & $522.0$ & $525.9$ & $2.9$ & 10/10 & $522.3$ & $524.7$ & $2.0$ & 10/10 & $521.8$ & $\mathbf{523.9}$ & $1.4$ & 10/10 & $522.3$ & $525.9$ & $2.6$ & 10/10 \\
a6-60 & $733.1$ & $738.8$ & $4.4$ & 10/10 & $724.6$ & $735.8$ & $8.0$ & 10/10 & $728.3$ & $734.1$ & $4.2$ & 10/10 & $725.4$ & $733.6$ & $5.7$ & 10/10 & $724.4$ & $\mathbf{731.3}$ & $4.8$ & 10/10 & $727.1$ & $732.5$ & $3.4$ & 10/10 \\
a6-72 & $840.1$ & $\mathbf{851.9}$ & $9.6$ & 9/10 & $841.1$ & $859.4$ & $12.5$ & 10/10 & $844.8$ & $852.9$ & $6.7$ & 10/10 & $840.2$ & $860.0$ & $11.1$ & 10/10 & $844.4$ & $860.8$ & $12.9$ & 10/10 & $845.8$ & $861.4$ & $8.8$ & 10/10 \\
a7-56 & $633.5$ & $637.2$ & $2.5$ & 10/10 & $631.7$ & $\mathbf{636.2}$ & $3.2$ & 10/10 & $633.2$ & $638.7$ & $3.1$ & 10/10 & $633.3$ & $637.4$ & $3.3$ & 10/10 & $634.5$ & $638.0$ & $3.1$ & 10/10 & $635.2$ & $638.2$ & $3.1$ & 10/10 \\
a7-70 & $804.8$ & $820.9$ & $11.5$ & 10/10 & $809.5$ & $818.9$ & $7.4$ & 10/10 & $806.4$ & $824.8$ & $11.6$ & 10/10 & $814.2$ & $823.2$ & $7.2$ & 10/10 & $807.7$ & $\mathbf{817.0}$ & $6.6$ & 10/10 & $810.0$ & $819.7$ & $6.0$ & 10/10 \\
a7-84 & $965.5$ & $980.5$ & $15.0$ & 10/10 & $952.3$ & $969.5$ & $9.3$ & 10/10 & $957.5$ & $983.7$ & $19.3$ & 10/10 & $954.0$ & $968.2$ & $12.9$ & 10/10 & $958.7$ & $969.9$ & $8.6$ & 10/10 & $952.6$ & $\mathbf{966.8}$ & $11.4$ & 10/10 \\
a8-64 & $671.2$ & $679.4$ & $5.4$ & 10/10 & $669.3$ & $681.4$ & $9.0$ & 10/10 & $669.1$ & $\mathbf{676.5}$ & $5.7$ & 10/10 & $668.7$ & $676.6$ & $5.5$ & 10/10 & $673.1$ & $679.3$ & $4.4$ & 10/10 & $672.0$ & $680.7$ & $7.6$ & 10/10 \\
a8-80 & $864.5$ & $875.0$ & $6.2$ & 10/10 & $860.4$ & $870.9$ & $7.4$ & 10/10 & $852.4$ & $873.2$ & $12.5$ & 10/10 & $855.0$ & $\mathbf{869.0}$ & $8.1$ & 10/10 & $873.7$ & $884.6$ & $10.2$ & 10/10 & $866.4$ & $877.8$ & $8.2$ & 10/10 \\
a8-96 & $1164.4$ & $1206.2$ & $27.3$ & 8/10 & $1170.3$ & $1197.1$ & $26.9$ & 10/10 & $1165.9$ & $1208.4$ & $35.6$ & 9/10 & $1144.1$ & $\mathbf{1181.0}$ & $24.3$ & 9/10 & $1167.6$ & $1210.9$ & $36.0$ & 9/10 & $1174.2$ & $1208.2$ & $30.6$ & 9/10 \\
\bottomrule
\end{tabular}
}
\end{table*}


\begin{table*}
\centering
\scriptsize
\setlength{\tabcolsep}{3pt}
\caption{Final best feasible objective values for Ropke instances for unlimited CS visits ($\gamma=0.7$).}
\label{tab:results_summary_unlimited_0.7}
\resizebox{\linewidth}{!}{%
\begin{tabular}{lrrrrrrrrrrrrrrrrrrrrrrrr}
\toprule
Instance & \multicolumn{4}{c}{BI-LNS-Greedy} & \multicolumn{4}{c}{BI-LNS-Regret} & \multicolumn{4}{c}{BI-LNS-Random} & \multicolumn{4}{c}{BI-LNS-Rule} & \multicolumn{4}{c}{BI-LNS-ALNS} & \multicolumn{4}{c}{BI-LNS-LLM} \\
 & Min & Mean & Std & Feas.  & Min & Mean & Std & Feas.  & Min & Mean & Std & Feas.  & Min & Mean & Std & Feas.  & Min & Mean & Std & Feas.  & Min & Mean & Std & Feas.  \\
\midrule
a5-60 & $770.5$ & $794.8$ & $30.6$ & 8/10 & $769.6$ & $782.3$ & $16.5$ & 8/10 & $753.1$ & $779.7$ & $15.0$ & 8/10 & $756.4$ & $783.5$ & $15.0$ & 9/10 & $755.9$ & $\mathbf{776.6}$ & $17.2$ & 10/10 & $770.3$ & $790.3$ & $21.5$ & 8/10 \\
a6-48 & $528.6$ & $534.6$ & $5.1$ & 10/10 & $527.6$ & $534.6$ & $4.8$ & 10/10 & $527.7$ & $535.2$ & $5.0$ & 10/10 & $529.8$ & $533.1$ & $3.2$ & 10/10 & $530.4$ & $537.0$ & $5.0$ & 10/10 & $527.6$ & $\mathbf{532.9}$ & $2.4$ & 10/10 \\
a6-60 & $738.4$ & $755.2$ & $15.2$ & 10/10 & $741.7$ & $757.3$ & $10.2$ & 10/10 & $743.1$ & $753.3$ & $11.0$ & 10/10 & $739.7$ & $754.2$ & $19.6$ & 10/10 & $743.2$ & $\mathbf{752.3}$ & $7.2$ & 10/10 & $741.6$ & $758.7$ & $15.1$ & 10/10 \\
a6-72 & $874.1$ & $888.7$ & $18.3$ & 5/10 & $875.2$ & $894.5$ & $16.5$ & 8/10 & $877.8$ & $899.5$ & $15.5$ & 10/10 & $859.8$ & $891.5$ & $19.6$ & 10/10 & $859.4$ & $901.9$ & $29.4$ & 9/10 & $867.6$ & $\mathbf{888.6}$ & $18.8$ & 9/10 \\
a7-56 & $647.4$ & $667.5$ & $13.7$ & 10/10 & $651.7$ & $661.2$ & $5.1$ & 10/10 & $646.0$ & $\mathbf{658.5}$ & $6.3$ & 10/10 & $649.0$ & $665.9$ & $13.2$ & 10/10 & $656.9$ & $667.2$ & $11.7$ & 10/10 & $653.8$ & $666.0$ & $8.2$ & 10/10 \\
a7-70 & $837.1$ & $851.2$ & $15.4$ & 10/10 & $835.3$ & $853.4$ & $13.4$ & 10/10 & $827.2$ & $847.5$ & $13.5$ & 10/10 & $833.4$ & $\mathbf{843.5}$ & $11.5$ & 10/10 & $832.3$ & $847.0$ & $10.0$ & 10/10 & $844.5$ & $868.1$ & $37.6$ & 10/10 \\
a7-84 & $970.3$ & $1030.1$ & $31.9$ & 8/10 & $974.3$ & $1005.7$ & $21.9$ & 10/10 & $983.3$ & $1035.0$ & $52.1$ & 7/10 & $968.1$ & $1004.9$ & $31.5$ & 9/10 & $982.5$ & $1008.9$ & $19.5$ & 8/10 & $966.6$ & $\mathbf{998.2}$ & $20.0$ & 7/10 \\
a8-64 & $681.3$ & $697.0$ & $11.6$ & 10/10 & $689.7$ & $695.7$ & $4.7$ & 10/10 & $683.9$ & $697.3$ & $7.6$ & 10/10 & $681.7$ & $696.1$ & $8.4$ & 10/10 & $685.7$ & $693.5$ & $6.1$ & 10/10 & $679.7$ & $\mathbf{691.4}$ & $10.6$ & 10/10 \\
a8-80 & $889.5$ & $917.5$ & $16.3$ & 8/10 & $890.8$ & $\mathbf{915.3}$ & $15.3$ & 9/10 & $891.5$ & $918.9$ & $25.6$ & 9/10 & $897.2$ & $931.8$ & $27.6$ & 9/10 & $880.5$ & $920.6$ & $22.7$ & 9/10 & $911.5$ & $924.7$ & $12.1$ & 8/10 \\
a8-96 & \text{NA} & \text{NA} & \text{NA} & 0/10 & $1229.1$ & $1261.3$ & $45.1$ & 3/10 & $1211.9$ & $1211.9$ & - & 1/10 & $1252.9$ & $1253.0$ & $0.2$ & 2/10 & $1278.0$ & $1278.0$ & - & 1/10 & $1203.1$ & $\mathbf{1203.1}$ & - & 1/10 \\
\bottomrule
\end{tabular}
}
\end{table*}
\begin{table*}
\centering
\scriptsize
\setlength{\tabcolsep}{3pt}
\caption{Final best feasible objective values for Ropke instances for unlimited CS visits ($\gamma=0.8$).}
\label{tab:results_summary_unlimited_0.8}
\resizebox{\linewidth}{!}{%
\begin{tabular}{lrrrrrrrrrrrrrrrrrrrrrrrr}
\toprule
Instance & \multicolumn{4}{c}{BI-LNS-Greedy} & \multicolumn{4}{c}{BI-LNS-Regret} & \multicolumn{4}{c}{BI-LNS-Random} & \multicolumn{4}{c}{BI-LNS-Rule} & \multicolumn{4}{c}{BI-LNS-ALNS} & \multicolumn{4}{c}{BI-LNS-LLM} \\
 & Min & Mean & Std & Feas.  & Min & Mean & Std & Feas.  & Min & Mean & Std & Feas.  & Min & Mean & Std & Feas.  & Min & Mean & Std & Feas.  & Min & Mean & Std & Feas.  \\
\midrule
a5-60 & \text{NA} & \text{NA} & \text{NA} & 0/10 & \text{NA} & \text{NA} & \text{NA} & 0/10 & $781.9$ & $798.7$ & $23.7$ & 2/10 & $781.9$ & $781.9$ & - & 1/10 & $778.1$ & $\mathbf{778.9}$ & $1.1$ & 2/10 & $801.3$ & $829.2$ & $39.5$ & 2/10 \\
a6-48 & $537.0$ & $548.1$ & $10.0$ & 10/10 & $533.3$ & $543.6$ & $7.5$ & 10/10 & $536.3$ & $544.1$ & $7.0$ & 10/10 & $535.3$ & $\mathbf{542.6}$ & $3.7$ & 10/10 & $535.7$ & $543.9$ & $6.0$ & 10/10 & $538.5$ & $552.0$ & $14.9$ & 10/10 \\
a6-60 & $754.7$ & $782.8$ & $25.4$ & 9/10 & $748.1$ & $765.0$ & $15.0$ & 10/10 & $747.8$ & $\mathbf{764.0}$ & $10.3$ & 10/10 & $750.7$ & $764.8$ & $9.0$ & 10/10 & $759.3$ & $775.4$ & $17.6$ & 10/10 & $749.5$ & $773.5$ & $20.7$ & 10/10 \\
a6-72 & $928.6$ & $929.7$ & $1.7$ & 2/10 & $864.0$ & $922.6$ & $48.2$ & 5/10 & $861.4$ & $\mathbf{910.8}$ & $32.4$ & 6/10 & $901.6$ & $914.1$ & $11.9$ & 3/10 & $900.2$ & $922.2$ & $19.2$ & 5/10 & $913.1$ & $920.9$ & $11.1$ & 2/10 \\
a7-56 & $685.4$ & $710.4$ & $19.0$ & 7/10 & $655.9$ & $692.9$ & $23.1$ & 8/10 & $670.9$ & $697.6$ & $18.3$ & 7/10 & $684.7$ & $712.9$ & $25.1$ & 5/10 & $673.7$ & $\mathbf{689.7}$ & $13.5$ & 7/10 & $682.6$ & $708.4$ & $16.0$ & 7/10 \\
a7-70 & $846.2$ & $877.7$ & $19.2$ & 10/10 & $845.3$ & $875.2$ & $21.0$ & 8/10 & $835.7$ & $\mathbf{866.4}$ & $29.7$ & 9/10 & $845.1$ & $868.6$ & $12.7$ & 8/10 & $832.0$ & $874.8$ & $40.4$ & 8/10 & $855.0$ & $881.6$ & $28.3$ & 4/10 \\
a7-84 & $1030.2$ & $1069.9$ & $25.3$ & 5/10 & $1037.2$ & $1054.6$ & $22.7$ & 3/10 & $1008.5$ & $\mathbf{1026.6}$ & $20.2$ & 4/10 & $990.1$ & $1029.9$ & $31.1$ & 5/10 & $1029.0$ & $1038.2$ & $10.2$ & 5/10 & $1005.7$ & $1027.8$ & $31.2$ & 2/10 \\
a8-64 & $695.2$ & $714.0$ & $17.8$ & 10/10 & $694.2$ & $\mathbf{708.4}$ & $10.1$ & 10/10 & $694.1$ & $711.4$ & $10.3$ & 10/10 & $702.0$ & $710.4$ & $7.6$ & 10/10 & $701.9$ & $710.8$ & $8.3$ & 10/10 & $696.3$ & $709.8$ & $11.5$ & 10/10 \\
a8-80 & $937.8$ & $952.3$ & $16.2$ & 4/10 & $972.8$ & $972.8$ & - & 1/10 & $940.8$ & $963.3$ & $28.5$ & 3/10 & $905.8$ & $\mathbf{940.6}$ & $30.8$ & 3/10 & $943.3$ & $954.0$ & $11.2$ & 4/10 & $937.3$ & $955.3$ & $24.7$ & 3/10 \\
a8-96 & \text{NA} & \text{NA} & \text{NA} & 0/10 & \text{NA} & \text{NA} & \text{NA} & 0/10 & $1280.2$ & $\mathbf{1280.2}$ & - & 1/10 & \text{NA} & \text{NA} & \text{NA} & 0/10 & \text{NA} & \text{NA} & \text{NA} & 0/10 & \text{NA} & \text{NA} & \text{NA} & 0/10 \\
\bottomrule
\end{tabular}
}
\end{table*}
\begin{table*}
\centering
\scriptsize
\setlength{\tabcolsep}{3pt}
\caption{Final best feasible objective values for Ropke instances for unlimited CS visits ($\gamma=0.9$).}
\label{tab:results_summary_unlimited_0.9}
\resizebox{\linewidth}{!}{%
\begin{tabular}{lrrrrrrrrrrrrrrrrrrrrrrrr}
\toprule
Instance & \multicolumn{4}{c}{BI-LNS-Greedy} & \multicolumn{4}{c}{BI-LNS-Regret} & \multicolumn{4}{c}{BI-LNS-Random} & \multicolumn{4}{c}{BI-LNS-Rule} & \multicolumn{4}{c}{BI-LNS-ALNS} & \multicolumn{4}{c}{BI-LNS-LLM} \\
 & Min & Mean & Std & Feas.  & Min & Mean & Std & Feas.  & Min & Mean & Std & Feas.  & Min & Mean & Std & Feas.  & Min & Mean & Std & Feas.  & Min & Mean & Std & Feas.  \\
\midrule
a5-60 & \text{NA} & \text{NA} & \text{NA} & 0/10 & \text{NA} & \text{NA} & \text{NA} & 0/10 & $801.5$ & $801.5$ & - & 1/10 & \text{NA} & \text{NA} & \text{NA} & 0/10 & \text{NA} & \text{NA} & \text{NA} & 0/10 & \text{NA} & \text{NA} & \text{NA} & 0/10 \\
a6-48 & $554.8$ & $567.2$ & $12.2$ & 6/10 & $551.5$ & $567.6$ & $11.5$ & 8/10 & $554.0$ & $566.2$ & $9.2$ & 9/10 & $551.4$ & $571.1$ & $20.7$ & 8/10 & $555.0$ & $570.4$ & $10.4$ & 8/10 & $551.9$ & $\mathbf{564.6}$ & $7.5$ & 8/10 \\
a6-60 & $779.7$ & $786.6$ & $6.8$ & 7/10 & $768.9$ & $\mathbf{780.7}$ & $8.9$ & 5/10 & $777.3$ & $792.7$ & $20.4$ & 6/10 & $768.0$ & $783.5$ & $12.2$ & 8/10 & $760.2$ & $787.1$ & $21.4$ & 10/10 & $763.4$ & $782.1$ & $20.6$ & 5/10 \\
a6-72 & \text{NA} & \text{NA} & \text{NA} & 0/10 & \text{NA} & \text{NA} & \text{NA} & 0/10 & $946.8$ & $946.8$ & - & 1/10 & \text{NA} & \text{NA} & \text{NA} & 0/10 & \text{NA} & \text{NA} & \text{NA} & 0/10 & \text{NA} & \text{NA} & \text{NA} & 0/10 \\
a7-56 & $737.6$ & $737.6$ & - & 1/10 & \text{NA} & \text{NA} & \text{NA} & 0/10 & $729.9$ & $729.9$ & - & 1/10 & \text{NA} & \text{NA} & \text{NA} & 0/10 & \text{NA} & \text{NA} & \text{NA} & 0/10 & \text{NA} & \text{NA} & \text{NA} & 0/10 \\
a7-70 & $890.7$ & $890.7$ & - & 1/10 & $862.9$ & $900.2$ & $42.2$ & 3/10 & $865.2$ & $897.0$ & $23.8$ & 5/10 & $893.6$ & $910.6$ & $17.6$ & 3/10 & $936.4$ & $946.0$ & $13.6$ & 2/10 & $884.8$ & $902.9$ & $17.1$ & 3/10 \\
a7-84 & \text{NA} & \text{NA} & \text{NA} & 0/10 & $1046.7$ & $1046.7$ & - & 1/10 & \text{NA} & \text{NA} & \text{NA} & 0/10 & $1063.7$ & $1063.7$ & - & 1/10 & \text{NA} & \text{NA} & \text{NA} & 0/10 & \text{NA} & \text{NA} & \text{NA} & 0/10 \\
a8-64 & $721.5$ & $\mathbf{732.7}$ & $10.0$ & 4/10 & $714.7$ & $737.6$ & $19.9$ & 7/10 & $701.5$ & $735.4$ & $19.4$ & 7/10 & $703.3$ & $749.6$ & $34.9$ & 6/10 & $717.9$ & $740.6$ & $17.5$ & 6/10 & $713.3$ & $736.9$ & $32.8$ & 4/10 \\
a8-80 & \text{NA} & \text{NA} & \text{NA} & 0/10 & \text{NA} & \text{NA} & \text{NA} & 0/10 & \text{NA} & \text{NA} & \text{NA} & 0/10 & \text{NA} & \text{NA} & \text{NA} & 0/10 & \text{NA} & \text{NA} & \text{NA} & 0/10 & \text{NA} & \text{NA} & \text{NA} & 0/10 \\
a8-96 & \text{NA} & \text{NA} & \text{NA} & 0/10 & \text{NA} & \text{NA} & \text{NA} & 0/10 & \text{NA} & \text{NA} & \text{NA} & 0/10 & \text{NA} & \text{NA} & \text{NA} & 0/10 & \text{NA} & \text{NA} & \text{NA} & 0/10 & \text{NA} & \text{NA} & \text{NA} & 0/10 \\
\bottomrule
\end{tabular}
}
\end{table*}

\end{landscape}

\newpage
\clearpage

\section{Results for the Chance-Constrained EADARP under CV=0.3}
\label{app_cc}

Table~\ref{tab:cc_results_unlimited_cv0.3} shows the results for the chance-constrained EADARP under $CV=0.3$ and $\gamma \in \{0.4, 0.7\}$. Results for higher uncertainty levels exhibit near-zero feasibility
at strict confidence levels across all methods.

\begin{table}[H]
\centering
\caption{Final best feasible objective values for the chance-constrained EADARP under $CV=0.3$ and $\gamma \in \{0.4, 0.7\}$.
For each method and confidence level $\alpha$, the mean and standard deviation
over 10 runs are reported together with the feasibility rate (feasible runs /
total runs). Best mean values per (instance, $\gamma$, $\alpha$) combination
are highlighted in bold.}
\label{tab:cc_results_unlimited_cv0.3}
\resizebox{\textwidth}{!}{
\begin{tabular}{llrrrrrrrrr|rrrrrrrrr}
\toprule
\multirow[c]{3}{*}{Instance} & \multirow[c]{3}{*}{$\alpha$} & \multicolumn{9}{c|}{$\gamma=0.4$} & \multicolumn{9}{c}{$\gamma=0.7$} \\
 &  & \multicolumn{3}{c}{BI-LNS-Greedy} & \multicolumn{3}{c}{BI-LNS-ALNS} & \multicolumn{3}{c|}{BI-LNS-LLM} & \multicolumn{3}{c}{BI-LNS-Greedy} & \multicolumn{3}{c}{BI-LNS-ALNS} & \multicolumn{3}{c}{BI-LNS-LLM} \\
 &  & Mean & Std & Feas. & Mean & Std & Feas. & Mean & Std & Feas. & Mean & Std & Feas. & Mean & Std & Feas. & Mean & Std & Feas. \\
\midrule
\multirow[c]{3}{*}{a3-18} & 0.99 & $\mathbf{263.2}$ & $0.0$ & 10/10 & $\mathbf{263.2}$ & $0.0$ & 10/10 & $\mathbf{263.2}$ & $0.0$ & 10/10 & $\mathbf{265.1}$ & $0.0$ & 10/10 & $\mathbf{265.1}$ & $0.0$ & 10/10 & $\mathbf{265.1}$ & $0.0$ & 10/10 \\
 & 0.9999 & - & - & 0/10 & - & - & 0/10 & - & - & 0/10 & - & - & 0/10 & - & - & 0/10 & - & - & 0/10 \\
 & 0.999999 & - & - & 0/10 & - & - & 0/10 & - & - & 0/10 & - & - & 0/10 & - & - & 0/10 & - & - & 0/10 \\
\cmidrule{1-20}
\multirow[c]{3}{*}{a4-24} & 0.99 & $\mathbf{342.4}$ & $0.0$ & 10/10 & $\mathbf{342.4}$ & $0.0$ & 10/10 & $\mathbf{342.4}$ & $0.0$ & 10/10 & $363.8$ & $6.4$ & 4/10 & $357.5$ & $5.0$ & 5/10 & $\mathbf{357.3}$ & $5.4$ & 3/10 \\
 & 0.9999 & $360.8$ & $5.2$ & 9/10 & $\mathbf{357.3}$ & $0.0$ & 10/10 & $\mathbf{357.3}$ & $0.0$ & 10/10 & $377.6$ & $4.1$ & 4/10 & $\mathbf{364.1}$ & $2.4$ & 2/10 & $372.8$ & $6.0$ & 5/10 \\
 & 0.999999 & - & - & 0/10 & - & - & 0/10 & - & - & 0/10 & - & - & 0/10 & - & - & 0/10 & - & - & 0/10 \\
\cmidrule{1-20}
\multirow[c]{3}{*}{a5-50} & 0.99 & $594.7$ & $4.7$ & 10/10 & $594.8$ & $4.8$ & 10/10 & $\mathbf{593.9}$ & $4.0$ & 10/10 & $661.0$ & $22.0$ & 3/10 & $\mathbf{653.9}$ & $9.7$ & 5/10 & $659.9$ & $33.5$ & 3/10 \\
 & 0.9999 & - & - & 0/10 & - & - & 0/10 & - & - & 0/10 & - & - & 0/10 & - & - & 0/10 & - & - & 0/10 \\
 & 0.999999 & - & - & 0/10 & - & - & 0/10 & - & - & 0/10 & - & - & 0/10 & - & - & 0/10 & - & - & 0/10 \\
\cmidrule{1-20}
\multirow[c]{3}{*}{a6-48} & 0.99 & $559.5$ & $9.0$ & 10/10 & $562.5$ & $5.3$ & 10/10 & $\mathbf{558.9}$ & $7.7$ & 10/10 & $594.6$ & $20.3$ & 10/10 & $588.3$ & $11.2$ & 10/10 & $\mathbf{584.3}$ & $7.4$ & 10/10 \\
 & 0.9999 & $583.1$ & $7.2$ & 10/10 & $583.3$ & $5.8$ & 10/10 & $\mathbf{580.2}$ & $5.6$ & 10/10 & $619.5$ & $7.9$ & 9/10 & $605.9$ & $8.4$ & 10/10 & $\mathbf{604.9}$ & $11.4$ & 9/10 \\
 & 0.999999 & - & - & 0/10 & - & - & 0/10 & - & - & 0/10 & - & - & 0/10 & - & - & 0/10 & - & - & 0/10 \\
\cmidrule{1-20}
\multirow[c]{3}{*}{a6-60} & 0.99 & - & - & 0/10 & - & - & 0/10 & - & - & 0/10 & - & - & 0/10 & - & - & 0/10 & - & - & 0/10 \\
 & 0.9999 & - & - & 0/10 & - & - & 0/10 & - & - & 0/10 & - & - & 0/10 & - & - & 0/10 & - & - & 0/10 \\
 & 0.999999 & - & - & 0/10 & - & - & 0/10 & - & - & 0/10 & - & - & 0/10 & - & - & 0/10 & - & - & 0/10 \\
\cmidrule{1-20}
\multirow[c]{3}{*}{a7-56} & 0.99 & $672.0$ & $5.1$ & 10/10 & $\mathbf{668.0}$ & $5.3$ & 10/10 & $670.0$ & $4.4$ & 10/10 & $722.2$ & $14.1$ & 6/10 & $723.6$ & $20.9$ & 8/10 & $\mathbf{720.7}$ & $19.9$ & 9/10 \\
 & 0.9999 & - & - & 0/10 & - & - & 0/10 & - & - & 0/10 & - & - & 0/10 & - & - & 0/10 & - & - & 0/10 \\
 & 0.999999 & - & - & 0/10 & - & - & 0/10 & - & - & 0/10 & - & - & 0/10 & - & - & 0/10 & - & - & 0/10 \\
\cmidrule{1-20}
\multirow[c]{3}{*}{a7-70} & 0.99 & $\mathbf{873.0}$ & $5.7$ & 10/10 & $874.6$ & $8.8$ & 10/10 & $875.5$ & $5.7$ & 10/10 & $953.4$ & $44.0$ & 2/10 & $\mathbf{930.4}$ & $15.3$ & 5/10 & $944.5$ & $28.3$ & 6/10 \\
 & 0.9999 & - & - & 0/10 & - & - & 0/10 & - & - & 0/10 & - & - & 0/10 & - & - & 0/10 & - & - & 0/10 \\
 & 0.999999 & - & - & 0/10 & - & - & 0/10 & - & - & 0/10 & - & - & 0/10 & - & - & 0/10 & - & - & 0/10 \\
\cmidrule{1-20}
\multirow[c]{3}{*}{a8-64} & 0.99 & $720.7$ & $7.2$ & 10/10 & $\mathbf{720.1}$ & $8.8$ & 10/10 & $723.3$ & $5.6$ & 10/10 & $757.1$ & $15.2$ & 8/10 & $\mathbf{752.7}$ & $11.6$ & 10/10 & $766.3$ & $20.1$ & 10/10 \\
 & 0.9999 & $724.1$ & $7.9$ & 10/10 & $\mathbf{721.3}$ & $7.0$ & 10/10 & $724.1$ & $7.9$ & 10/10 & $766.0$ & $15.6$ & 10/10 & $\mathbf{765.4}$ & $19.2$ & 10/10 & $768.9$ & $20.4$ & 9/10 \\
 & 0.999999 & - & - & 0/10 & - & - & 0/10 & - & - & 0/10 & - & - & 0/10 & - & - & 0/10 & - & - & 0/10 \\
\bottomrule
\end{tabular}
}
\end{table}

\end{document}